\documentclass{article}

\usepackage{microtype}
\usepackage{graphicx}
\usepackage{subcaption}
\usepackage{booktabs} 

\usepackage{hyperref}

\usepackage[final,main]{neurips_2026}

\usepackage{hyperref}
\usepackage{url}
\usepackage{graphicx}

\usepackage{multicol}
\usepackage{multirow}
\usepackage{graphicx} 
\usepackage{booktabs} 
\usepackage{xcolor}   
\usepackage{pifont}   
\usepackage{array}
\definecolor{ForestGreen}{RGB}{34,139,34} 
\newcommand{\cmark}{\ding{51}}%
\newcommand{\xmark}{\ding{55}}%
\usepackage{siunitx} 
\usepackage{makecell}
\usepackage{wrapfig}
\usepackage{cleveref}
\definecolor{ForestGreen}{RGB}{34,139,34}

\usepackage[textsize=tiny]{todonotes}

\title{Beyond Bag-of-Words: Diagnosing Compositional Binding Failures in Vision-Language Models}

\author{
  Cristian Sbrolli \\
  Politecnico di Milan, AIRLab\\
  \texttt{cristian.sbrolli@polimi.it} \\
  \And
  Toshihiko Yamasaki \\
  The University of Tokyo, Computer Vision and Media Lab\\
  \texttt{yamasaki@cvm.t.u-tokyo.ac.jp}
  \And
  Matteo Matteucci \\
  Politecnico di Milano, AIRLab \\
  \texttt{matteo.matteucci@polimi.it} \\
}

\begin{document}

\maketitle

\begin{abstract}
Modern vision-language models struggle with basic compositional reasoning, failing to bind attributes to objects or relations to their referents. Existing benchmarks either rely on noisy real images that conflate confounding visual variables with the reasoning failure, or use simplistic synthetic scenes lacking the realism modern VLMs are tuned for. We introduce \textbf{Auto-Comp}, a fully automated, concept-driven pipeline that bridges this gap by generating photorealistic compositional benchmarks at scale. Its core innovation is a \textit{parallel A/B construction}: for each concept, the pipeline emits a \textit{Minimal} sample (template caption, isolated objects on a white background) and a \textit{Contextual} sample (LLM-rewritten caption, objects embedded in a realistic scene), isolating core binding ability from visio-linguistic complexity. We instantiate \textit{four} task families spanning the two canonical axes of compositional binding: \textit{Color} and \textit{Shape-Color} (attribute binding), and \textit{Position} and \textit{Relative Size} (relational binding). We evaluate over 25 VLMs spanning CLIP, SigLIP, hard-negative-trained, and frontier generative models. The findings are consistent across architectures and scales: every model exhibits a large Swap-vs-Confusion gap, with low-entropy distractors (e.g., repeated objects or colors) exposing failures \textit{beyond} the known bag-of-words limitations. We further uncover a task-dependent trade-off: visio-linguistic context aids relational reasoning but hinders attribute binding through visual clutter. We publicly release the pipeline and benchmarks.
\end{abstract}

\section{Introduction}
\label{sec:intro} 
Vision-Language Models (VLMs), from contrastive dual-encoders such as CLIP and SigLIP to frontier multi-modal LLMs (MLLMs), have demonstrated remarkable capabilities in connecting text and images, enabling applications ranging from detailed image captioning to segmentation and image generation~\citep{radford2021learning, ramesh2022hierarchical, rao2022denseclip}. Yet a critical failure point persists across all of them: \textit{compositional reasoning}, the ability to correctly bind elements within a scene based on their attributes and relationships. This failure is not a minor flaw: it reveals a fundamental weakness in scene understanding and limits the reliability of VLMs in high-stakes applications.

\begin{table}[t]
\small
\centering
\caption{Comparison of compositional VLM benchmarks. Auto-Comp is the first to introduce a paired A/B context construction, built on a fully automatic, concept-driven pipeline with photorealistic images and text. The resulting suite spans four task families across attribute and relational binding.}
\label{tab:benchmark_comparison}
\resizebox{\textwidth}{!}{%
\begin{tabular}{l|c|ccccc}
\toprule
\textbf{Benchmark} & \textbf{\# of Samples} & \textbf{Realistic Images} & \textbf{Realistic Text} & \textbf{Concept-driven Generation} & \textbf{Fully Automatic} & \textbf{Context A/B}\\
\midrule
CREPE~\citep{ma2023crepe} & 370,000+ & \textcolor{ForestGreen}{\cmark} & \textcolor{red}{\xmark} & \textcolor{red}{\xmark} & \textcolor{ForestGreen}{\cmark} & \textcolor{red}{\xmark}\\
SugarCrepe~\citep{hsieh2023sugarcrepe} & 7,512 & \textcolor{ForestGreen}{\cmark} & \textcolor{ForestGreen}{\cmark} & \textcolor{red}{\xmark} & \textcolor{red}{\xmark}& \textcolor{red}{\xmark} \\
SugarCrepe++~\citep{dumpala2024sugarcrepe++} & 4,757 & \textcolor{ForestGreen}{\cmark} & \textcolor{ForestGreen}{\cmark} & \textcolor{red}{\xmark} & \textcolor{red}{\xmark}& \textcolor{red}{\xmark}\\
Winoground~\citep{thrush2022winoground} & 400 & \textcolor{ForestGreen}{\cmark} & \textcolor{ForestGreen}{\cmark} & \textcolor{red}{\xmark} & \textcolor{red}{\xmark} & \textcolor{red}{\xmark}\\
SPEC~\citep{peng2024spec} & $\sim$6000 & \textcolor{ForestGreen}{\cmark} & \textcolor{red}{\xmark} & \textcolor{ForestGreen}{\cmark} & \textcolor{ForestGreen}{\cmark} & \textcolor{red}{\xmark} \\
CIVET~\citep{rizzoli2025civet} & $\sim$35,000 & \textcolor{red}{\xmark} & \textcolor{ForestGreen}{\cmark} & \textcolor{ForestGreen}{\cmark} & \textcolor{ForestGreen}{\cmark} & \textcolor{red}{\xmark}\\
\midrule
\textbf{Auto-Comp Suite (Ours)} & \textbf{200,000+} & \textcolor{ForestGreen}{\cmark} & \textcolor{ForestGreen}{\cmark} & \textcolor{ForestGreen}{\cmark} & \textcolor{ForestGreen}{\cmark} & \textcolor{ForestGreen}{\cmark} \\
\bottomrule
\end{tabular}%
}
\vspace{-1em}
\end{table}

Diagnosing this failure rigorously is difficult along two distinct axes. The first concerns \textit{what} to measure. Prior work has relied predominantly on swap-style hard negatives, where attributes or objects are permuted to expose bag-of-words behavior in dual-encoder VLMs. This lens is increasingly inadequate for the current model landscape: contrastive backbones such as CLIP and SigLIP, still relevant as they are widely used as feature extractors in downstream systems, continue to struggle on swap negatives, but modern MLLMs make substantial progress on simpler compositional configurations. Swap accuracy alone therefore may no longer significantly discriminate the field's strongest models. The second axis concerns \textit{how} to measure. Benchmarks built on real-world images~\citep{krishna2017visual,ma2023crepe,hsieh2023sugarcrepe,dumpala2024sugarcrepe++}, while ecologically valid, are inherently noisy,  making it impossible to isolate specific reasoning failures from confounding visual variables. Synthetic benchmarks offer more control but typically lack photorealism or use template-based language that fails to mirror real-world linguistic diversity~\citep{rizzoli2025civet}. This
leaves a critical question unanswered: do these models have a compositional ability that is simply brittle to the complexity of real scenes, or is their understanding fundamentally flawed? A diagnostic tool that addresses both axes simultaneously is therefore missing.

To fill this gap, we introduce \textbf{Auto-Comp}, a fully automated, concept-driven pipeline for generating scalable, photorealistic compositional benchmarks. On the diagnostic axis, alongside standard Swap negatives, we introduce the \textit{Confusion} benchmark, a stronger probe built from \textit{low-entropy distractors}: captions that recombine the same vocabulary present in the image with replacement (e.g., ``a red cube and a red sphere'' when the scene contains a red cube and a blue sphere). Such distractors test whether models truly bind attributes to their referents or rely on brittle co-occurrence heuristics, and as we show, they remain harder even for larger MLLMs. On the data axis, the core innovation is a \textit{parallel A/B} generation process: for each concept, Auto-Comp emits two visio-linguistic conditions, a \textit{Minimal} condition with a template caption and objects isolated on a white background, and a \textit{Contextual} condition where an LLM rewrites the caption into natural language and a text-to-image model renders the objects within a realistic scene. Pairing them enables a controlled test of how performance changes when moving from a sterile to a complex environment. The pipeline is \textit{general}: by swapping vocabularies, it can generate benchmarks for any user-defined compositional skill. We instantiate four such skills covering both canonical axes of compositional binding, \textit{attribute binding} (Color, Shape) and \textit{relational binding} (Position, Relative Size), and validate every sample with a two-stage automated pipeline audited against human annotators. \\
We make the following contributions:

\noindent\textbf{(i)} We present \textbf{Auto-Comp}, a fully automated and scalable pipeline for generating and validating high-quality, photorealistic synthetic datasets, enabling the creation of custom evaluation data for any user-defined compositional skill.

\noindent\textbf{(ii)} We propose a novel \textbf{A/B evaluation framework} using parallel \textit{Minimal} and \textit{Contextual} conditions that isolates core visual binding failures from visio-linguistic contextual complexity, and we introduce a novel \textit{Confusion Benchmark} that exposes a common vulnerability to low-entropy distractors, a failure mode going beyond the known ``bag-of-words'' issue identified by previous studies.

\noindent\textbf{(iii)} Using Auto-Comp, we generate the \textbf{Auto-Comp Suite}, comprising four large-scale benchmarks across attribute and relational binding, each at compositional complexities N=2 and N=3 (where N is the number of objects in a scene), with both \textit{Swap} and the novel \textit{Confusion} hard-negative schemes.

\noindent\textbf{(iv)} We conduct an extensive analysis of \textbf{25+ VLMs}, spanning CLIP and SigLIP architectures, hard-negative-trained, and frontier MLLMs up to 122B parameters. While modern MLLMs largely close the Swap gap on simpler configurations, the Confusion failure mode is universal: every model leaves large Swap-vs-Confusion gaps. We also uncover a task-dependent trade-off where visio-linguistic context aids relational reasoning but hinders attribute binding.

\section{Related Work}

Our research is situated at the intersection of VLM evaluation, compositional reasoning, and synthetic data generation. We position our contributions in relation to the key paradigms of prior work, summarized in~\Cref{tab:benchmark_comparison}.

\noindent\textbf{Data-driven Benchmarks on Real Images.}
A primary approach to compositional VLM evaluation uses pre-existing datasets, where innovation focuses on curating the negative texts. The large-scale CREPE benchmark~\citep{ma2023crepe}, for instance, starts from real image-caption pairs and programmatically swaps attributes to create hard negatives. While automated, this process can produce linguistically awkward captions, creating an artifact that models may exploit. Addressing this, SugarCrepe~\citep{hsieh2023sugarcrepe} and SugarCrepe++~\citep{dumpala2024sugarcrepe++}, leverage LLMs to generate fluent, realistic negative captions, significantly improving text quality. However, despite achieving realism in both modalities, these benchmarks require human evaluation and remain fundamentally data-driven, not concept-driven. Related linguistic-phenomena probes such as VALSE~\citep{parcalabescu2022valse}, ARO~\citep{yuksekgonuland}, and Whatsup~\citep{kamath2023s} construct minimal-edit foils on real images, isolating bag-of-words and spatial-relation failures, but inherit the same constraint: they are limited by what pre-existing images contain and cannot control distracting elements.

\noindent\textbf{Concept-driven Benchmarks with Synthetic Visuals.}
To gain composition control, a second line of research uses synthetic visuals. CIVET~\citep{rizzoli2025civet} programmatically places real-object cutouts onto a simple grid, offering precise concept-level control at the cost of photorealism and creating a domain gap with the organic images used to train modern VLMs. SPEC~\citep{peng2024spec} narrows this gap by rendering images for compositional probing, but at small scale ($\sim$6K items), with programmatic rather than naturalistic captions, single-edit foils, and without an A/B construction isolating core binding from visio-linguistic complexity.

\noindent\textbf{Unifying Realism and Concept-driven Generation.}
Our work pushes this combination substantially further along scale, A/B design, and validation rigor. We introduce \textbf{Auto-Comp}, a framework that pairs the concept-driven control of synthetic benchmarks with the photorealism of data-driven approaches. As summarized in~\Cref{tab:benchmark_comparison}, our pipeline fully automatically generates high-fidelity images and realistic text from any user-defined concept. Crucially, our framework is the first to introduce the parallel generation of \textit{Minimal} and \textit{Contextual} conditions for each concept. This construction provides a unique diagnostic tool for measuring the performance gap between a model's foundational binding capacity in sterile environments and its practical robustness in complex scenes.

\section{Methodology}
\label{sec:methodology}
\begin{figure}[t]
    \centering
    \includegraphics[width=\linewidth]{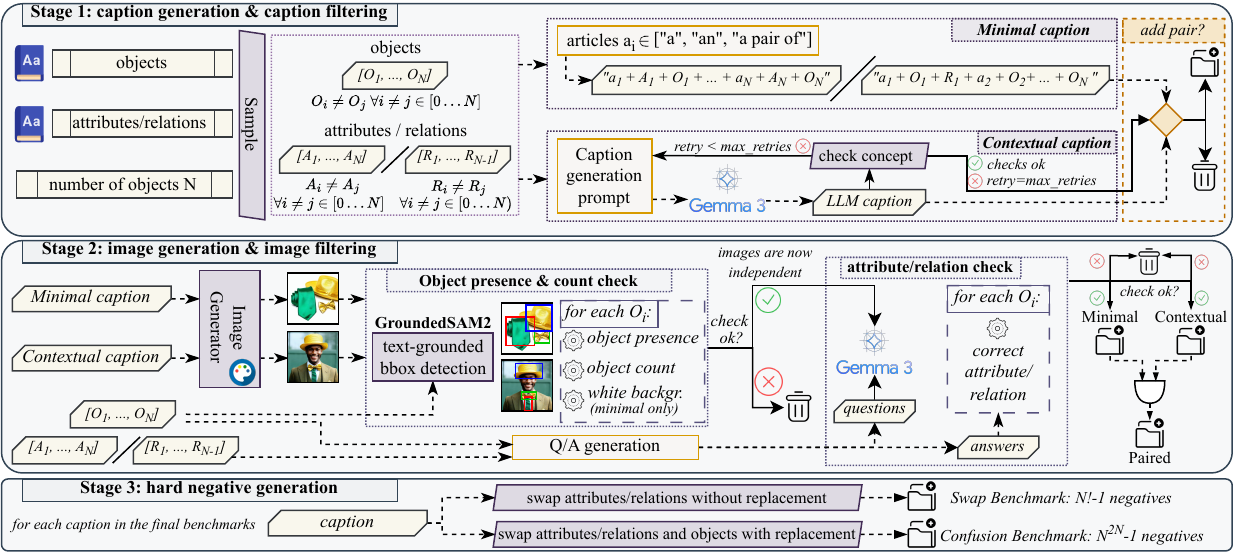}
    \caption{The Auto-Comp pipeline for automated benchmark creation. (1) Caption Generation: Concept-driven generation of template-based Minimal and LLM-based Contextual captions. (2) Image Synthesis \& Validation: A text-to-image model generates visuals, which are then validated using GroundedSAM2 and VLM checks. (3) Hard Negative Generation: Validated pairs are systematically altered by swapping attributes to create challenging compositional test sets.}
    \label{fig:autocompGA}
\end{figure}

As in~\Cref{fig:autocompGA}, the process consists of three main stages.

\noindent\textbf{Concept Definition.}
The foundation of our pipeline is the \textit{Concept}, a structured definition of a scene's intended ground truth. A concept $C$ is formally defined as a tuple $C = (\mathcal{O}_C, \mathcal{A}_C)$, where $\mathcal{O}_C = \{o_1, \dots, o_N\}$ is a set of $N$ objects or shapes, and $\mathcal{A}_C$ is a set of attributes. The structure of $\mathcal{A}_C$ is task-dependent and instantiates either \textit{attribute binding} or \textit{relational binding}: for \textit{attribute binding}, $\mathcal{A}_C$ contains $N$ attributes with a one-to-one mapping to ${O}_C$; for \textit{Position Binding} and \textit{Relative Size}, it contains $N-1$ binary relations that chain the objects in sequence. The parameter $N$ controls compositional complexity, with $N=1$ serving as a baseline and $N \ge 2$ used to evaluate true compositional binding. Vocabularies for all four task families are detailed in~\Cref{sec:benchmarks}.

\noindent\textbf{Stage 1: Parallel Caption Generation.}
For each concept, our pipeline generates two distinct styles of captions in parallel. This dual-track generation, designed to represent the same ground-truth with different visio-linguistic properties, is the foundation of our A/B testing framework.

\textit{Minimal Captions.} This track produces a direct, structured description of the concept using a template-based approach. The template joins the object-attribute pairs into a factual list, handling grammatical nuances (e.g., articles, pluralization) and adding the suffix "on a white background". The resulting caption serves to test a model's core binding ability in an isolated, unambiguous setting.

\textit{Contextual Captions.} This track produces a fluent, naturalistic description of the same concept. The concept is provided to a generative LLM (Gemma3-12b-it~\citep{gemmateam2025gemma3technicalreport}) with a few-shot prompt that defines its role and provides instructions and examples. The full prompt is provided in the Appendix~\ref{app:prompt}. To guarantee the LLM's adherence to these instructions, we perform an automated \textit{semantic preservation check} immediately after generation. This check programmatically parses the LLM's output to verify that all original objects and their assigned attributes from the concept $C$ are present and correctly associated. We define a "correct association" with a strict syntactic pattern:

    \textbf{For Attribute Binding (Color, Shape-Color),} given an
    object-attribute pair $(o_i, a_i)$, the generated text must contain a
    substring that matches:
    \begin{center}
    $\langle \texttt{Article} \rangle \enspace \langle \texttt{Word} \rangle \enspace a_i \enspace o_i$
    \end{center}
    This pattern requires the attribute $a_i$ to precede the object $o_i$,
    optionally separated by a modifier word.

    \textbf{For Relational Binding (Position, Relative Size),} given a chained relation $(o_i, r_i, o_{i+1})$, the text must match the pattern:
    \begin{center}
    $\langle \texttt{Article} \rangle \enspace [\langle \texttt{Word} \rangle] \enspace o_i \enspace r_i \enspace \langle \texttt{Article} \rangle \enspace [\langle \texttt{Word} \rangle] \enspace o_{i+1}$
    \end{center}
    This ensures the relation $r_k$ correctly links the objects, allowing for an optional modifier before each.

If the LLM's output fails to match these patterns for all required bindings after several retries, the entire concept is discarded. This means both the failed Contextual Caption and its successfully generated Minimal Caption are removed. This strict pairing protocol guarantees a clean, one-to-one correspondence for every concept that proceeds to the image generation stage.

\noindent\textbf{Stage 2: Image Synthesis and Automated Validation.}
From this point forward, the Minimal and Contextual tracks are processed independently to maximize the final dataset sizes. For each caption, we generate an image using StableDiffusion3.5-large~\citep{sd35} (model choices are ablated in Appendix~\ref{app:modelsel}). Recognizing that generative models themselves can fail at composition, we introduce a rigorous, two-stage validation pipeline for every generated image to ensure its fidelity.

\textit{Object Presence and Count Check.} The first, faster validation stage verifies that all objects from the source concept $C$ are present in the image in their correct number. We employ GroundedSAM2~\citep{ren2024grounded,ravi2024sam2segmentimages}, an open-vocabulary detection model, querying it with each object's class name to handle both singular and plural nouns (e.g., ensuring "gloves" corresponds to two instances). Formally, this validation, $V_{\text{obj}}(G(c), C)$, passes only if the detected objects perfectly match the concept's objects $\mathcal{O}_C$ in both class and cardinality.

\textit{Background Check for Minimal Images.} For images in the Minimal track, we perform an additional validation step. Using the bounding boxes from the object check, we mask all objects and analyze the remaining pixels to verify the background is uniformly white. This check, detailed in Appendix~\ref{app:implementationhyperparams}, allows for minor lighting variations and shadows inherent to the generative process, ensuring each image adheres to the "on a white background" clause and provides a truly sterile visual environment.

\textit{Semantic Correctness Check.} For an image $I \in \mathcal{I}'_{\text{min}} \cup \mathcal{I}'_{\text{ctx}}$ that has passed the object check, we next verify its attributes. We programmatically generate direct questions based on the source concept $C$ (e.g., "\textit{What color is the monitor?}") and leverage the Gemma-3-12b-it model~\citep{gemmateam2025gemma3technicalreport} to provide the answer. To prevent ambiguity, we constrain the VLM's response to our predefined attribute vocabularies. Formally, we define this validation function as $V_{\text{attr}}(I, C)$, which returns true only if the VLM provides the correct answer for all attribute questions derived from $C$. 

To ensure the reliability of the proposed image-validation approach, we first conducted a comprehensive human evaluation study (\Cref{sec:ablation}) to validate our automated judge. This study established that our VLM achieves over 94\% concordance with expert human annotators and is further confirmed by a cross-validator agreement study against a stronger judge (Qwen2.5-VL-72B, Appendix~\ref{app:validator_robustness}).

\begin{figure}[t]
    \centering
    \includegraphics[width=0.99\linewidth]{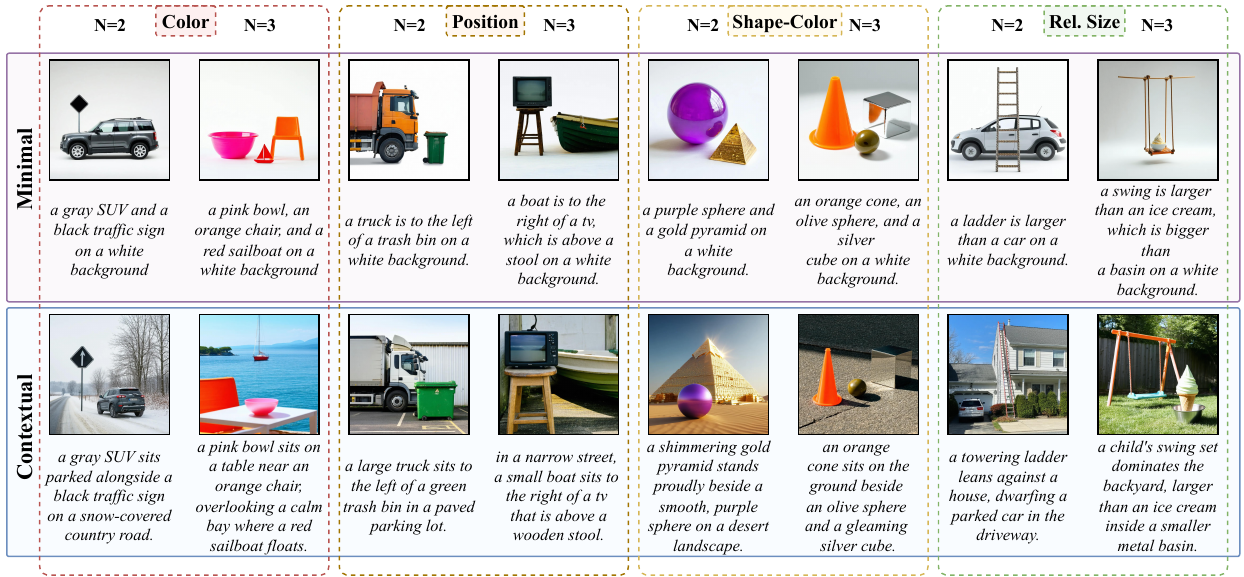}
    \caption{Examples of positive caption-image pairs from the four Auto-Comp benchmarks at $N=2$ and $N=3$. Each column shows one task family; rows contrast the \textit{Minimal} (white-background) and \textit{Contextual} (realistic-scene) renderings of the same concept, illustrating the A/B construction. }
    \label{fig:examples}
\end{figure}

\textit{Positive Benchmark.} The positive benchmarks are curated from samples that successfully pass both validation stages. Our large, standalone benchmarks, $\mathcal{B}_{\text{min}}$ and $\mathcal{B}_{\text{ctx}}$, are composed of all validated image-caption pairs for the Minimal and Contextual tracks, respectively. Finally, our \textit{Paired Comparison Set}, $\mathcal{B}_{\text{paired}}$, is created by taking the intersection of these two sets based on their source concepts. This set contains only concepts for which \textit{both} the Minimal and Contextual versions were successfully validated, enabling a perfectly controlled A/B test. We show qualitative examples in \Cref{fig:examples}.

\noindent\textbf{Stage 3: Hard Negative Benchmark Generation}
The final stage of our pipeline programmatically generates hard negative captions. Unlike prior work that modifies complex, real-world captions, our pipeline begins with text whose structure has been programmatically generated and verified. This high degree of control is a key strength. Unlike LLM-based rewriting, which can inadvertently alter sentence structure, our programmatic swapping ensures positive and negative captions are perfectly controlled, identical in length and complexity, differing only in the swapped conceptual elements. Our pipeline automatically handles all grammatical details to maintain linguistic naturalness. As we will demonstrate with a blind LLM evaluation (Appendix~\ref{app:blind}), this process creates negatives that are free of artifacts, forcing models to reason compositionally. For each concept with $N \ge 2$ objects, we generate two distinct hard negative benchmarks:

\textit{The Swap Benchmark.} This benchmark is a direct test of compositional binding. Our framework handles two general cases for creating swap-based negatives. For \textit{Attribute Binding Tasks} involving properties like color, we generate negatives by permuting the $N$ attributes among the $N$ fixed objects. For \textit{Relational Binding Tasks} (e.g., spatial position), where permuting the relations is nonsensical for $N=2$, we instead create foils by permuting the $N$ objects themselves (e.g., "\textit{a table on top of a chair}" from "\textit{a chair on top of a table}"). In either case, this process generates $N!-1$ unique hard negatives, directly measuring a model's ability to associate each element with its correct role.

\textit{The Confusion Benchmark.} To move beyond simple binding errors, we design the Confusion Benchmark, a more strenuous test intended to probe for reliance on brittle heuristics and susceptibility to low-entropy distractors. Here, we generate an exhaustive set of captions by creating all possible combinations of objects and attributes from the original concept, \textit{with replacement}. The generation process depends on the attribute type. For an attribute like color, where each of the $N$ objects is assigned one of $N$ colors, we generate all $N^N$ object arrangements and all $N^N$ color arrangements. The Cartesian product yields $N^{2N}-1$ hard-negative captions. For position, which describes the $N-1$ relative relations between the $N$ objects, the calculation is different. We again generate all $N^N$ object arrangements. However, we generate all possible arrangements for the $N-1$ relation slots, sampling from the $N-1$ relation types present in the caption. This results in $(N-1)^{N-1}$ unique positional arrangements, producing a total of $N^N(N-1)^{N-1} - 1$ hard negatives. This benchmark creates a massive set of challenging distractors, allowing us to quantify a model's robustness. 

\textit{Evaluation Protocol.} For each (image, candidate-set) pair, both contrastive and generative VLMs select the candidate caption with the highest score under their native scoring function, and accuracy is reported as the rate at which the positive caption is selected. For contrastive backbones (CLIP, OpenCLIP, SigLIP, SigLIP2), we use the standard image-text similarity score: cosine similarity for CLIP-family models and the sigmoid-loss compatible score of~\citep{zhai2023sigmoid} for SigLIP/SigLIP2. For generative MLLMs, we score each candidate caption by its log-likelihood under the model conditioned on the image, summed over the caption's tokens and obtained from a single forward pass per candidate. Both protocols are deterministic and produce a single, reproducible accuracy score. As all candidates are scored independently, the full negative set is used for every model regardless of its size.

\section{The Auto-Comp Benchmark Suite}
\label{sec:benchmarks}
The Auto-Comp suite relies on five carefully curated vocabularies:

\noindent\textit{Object Vocabulary.} Curated from the 365 classes of the Object365 dataset~\citep{shao2019objects365}, filtering ambiguous concepts, abstract nouns, and non-object categories (e.g., ``person,'' animals) to avoid common model biases, yielding 250 distinct objects.

\noindent\textit{Color Vocabulary.} 20 common colors built upon standard VGA colors (e.g., red, blue, green, yellow, magenta) and supplemented with common shades (e.g., gold, brown) to create challenging samples with similar yet distinct colors.

\noindent\textit{Spatial Relation Vocabulary.} 4 fundamental binary spatial relations inspired by prior work~\citep{rizzoli2025civet}: \textit{above}, \textit{under}, \textit{to the left of}, \textit{to the right of}.

\noindent\textit{Shape Vocabulary.} A set of 3D geometric primitives (e.g., \textit{cube}, \textit{sphere}, \textit{pyramid}, \textit{cone}, \textit{cylinder}). Unlike the Object Vocabulary, where real-object identities carry strong co-occurrence priors with specific colors that can shortcut compositional reasoning, abstract primitives are prior-free with respect to color: any color is equally plausible for any shape.

\noindent\textit{Size-Relation Vocabulary.} A set of comparative size descriptors (\textit{larger than}, \textit{smaller than}, \textit{bigger than}) probing comparative relational reasoning.

We instantiate the Auto-Comp pipeline on four benchmarks, chosen to cover the two canonical axes of compositional binding, attribute binding (Color, Shape-Color) and relational binding (Position, Relative Size), with two complexity levels each (N=2 and N=3). Color, Position, and Relative Size use the vocabulary of real objects and thus test binding under naturalistic priors; Shape-Color uses abstract geometric primitives, isolating compositional binding from learned object-attribute statistics.

\noindent
\textbf{Dataset Statistics and Analysis.}
We generate $100{,}000$ pairs of Minimal/Contextual captions per benchmark in Stage~1. After validation and hard-negative construction, Auto-Comp yields a large-scale benchmark suite across the four main task families at $N{=}2$ and $N{=}3$ (\Cref{tab:dataset_stats}). Producing this suite is computationally efficient: a batch of 100k concepts takes ~9.6 hours on 16 A100s (Appendix~\ref{app:efficiency_cost}). The validation survival rates (\Cref{tab:survival_rates}) expose where generators fail most: the \textit{Object Presence Check} filters many samples, the \textit{Background Check} further purifies the Minimal set, and the largest drop typically occurs at the final \textit{Attribute/Relation Correctness Check}. We additionally generate a Color $N{=}1$ split as a sanity check, reported in Appendix~\ref{app:sanity_checks}. To quantify how far the retained label distribution departs from the target prior, \Cref{tab:filter_bias} reports a summary of filter-induced distribution shift on the hardest $N{=}3$ setting, averaged over Minimal and Contextual.

\begin{table*}[t]
\centering

\begin{minipage}[t]{0.28\textwidth}
\centering
\scriptsize
\setlength{\tabcolsep}{3pt}
\captionof{table}{Final number of validated samples on all four benchmarks.}
\label{tab:dataset_stats}
\begin{tabular}{lc|ccc}
\toprule
\textbf{Task} & \textbf{N} & \textbf{Min} & \textbf{Ctx} & \textbf{Pair} \\
\midrule
Color      & 2 & 26,388 & 36,200 & 15,274 \\
Color      & 3 &  3,074 &  2,930 &  1,027 \\
Pos.       & 2 & 38,393 & 30,694 & 17,345 \\
Pos.       & 3 &  7,478 &  5,234 &  1,201 \\
Shape      & 2 & 18,642 & 15,388 &  8,214 \\
Shape      & 3 &  4,286 &  3,761 &  1,144 \\
Rel. Size  & 2 & 31,574 & 24,912 & 13,406 \\
Rel. Size  & 3 &  6,118 &  4,507 &  1,063 \\
\bottomrule
\end{tabular}
\end{minipage}
\hfill
\begin{minipage}[t]{0.37\textwidth}
\centering
\scriptsize
\setlength{\tabcolsep}{3pt}
\captionof{table}{Per benchmark survival rates (\%) after each validation stage.}
\label{tab:survival_rates}
\resizebox{\linewidth}{!}{%
\begin{tabular}{lc|ccc|cc}
\toprule
& & \multicolumn{3}{c|}{\textbf{Min}} & \multicolumn{2}{c}{\textbf{Ctx}} \\
\textbf{Task} & \textbf{N} & \textbf{Obj.} & \textbf{BG} & \textbf{Attr.} & \textbf{Obj.} & \textbf{Attr.} \\
\midrule
Color      & 2 & 74.1 & 67.6 & 26.4 & 79.5 & 36.2 \\
Color      & 3 & 33.6 & 29.1 &  3.1 & 35.7 &  2.9 \\
Pos.       & 2 & 78.8 & 65.2 & 38.4 & 75.4 & 30.7 \\
Pos.       & 3 & 49.2 & 31.6 &  7.5 & 43.3 &  5.2 \\
Shape      & 2 & 69.4 & 61.8 & 28.6 & 77.1 & 41.4 \\
Shape      & 3 & 39.7 & 31.5 &  4.3 & 37.9 &  3.8 \\
Rel. Size  & 2 & 72.4 & 58.7 & 31.6 & 68.9 & 24.9 \\
Rel. Size  & 3 & 41.8 & 26.9 &  6.1 & 39.5 &  4.5 \\
\bottomrule
\end{tabular}%
}
\end{minipage}
\hfill
\begin{minipage}[t]{0.25\textwidth}
\centering
\scriptsize
\setlength{\tabcolsep}{3pt}
\captionof{table}{Filter-shift. 
\textbf{In} = \% labels within $0.5{-}1.5\times$ target prior; 
\\ \textbf{Low} = \% labels below $0.25\times$ target prior; 
\\ \textbf{TV} = total-variation distance from target prior.}
\label{tab:filter_bias}
\vspace{-0.8em}
\resizebox{\linewidth}{!}{%
\begin{tabular}{lccc}
\toprule
\textbf{Task} & \textbf{In} & \textbf{Low} & \textbf{TV} \\
\midrule
Color      & 76.3 & 18.5 & 0.214 \\
Pos.       & 70.0 & 13.9 & 0.193 \\
Shape      & 67.7 & 17.4 & 0.208 \\
Rel. Size  & 73.1 & 11.7 & 0.179 \\
\bottomrule
\end{tabular}
}
\end{minipage}
\end{table*}

\begin{table*}[t]
\tiny
\centering
\setlength{\tabcolsep}{3.1pt}
\caption{Main results on the four Auto-Comp benchmarks. Each entry reports Swap / Confusion accuracy (\%). Random-chance baselines for Swap are 50.0\% / 16.7\% at N=2 / N=3; for Confusion, random chance baselines are 6.3\% / 0.14\% on attribute tasks and 25.0\% / 0.93\% on relational tasks.}
\label{tab:main_results}
\resizebox{\textwidth}{!}{%
\begin{tabular}{lll|cccccccccc}
\toprule
\textbf{Benchmark} & $\mathbf{N}$ & \textbf{Cond.}
& \makecell{\textbf{CLIP}\\B/16}
& \makecell{\textbf{CLIP}\\H/14-378}
& \makecell{\textbf{SigLIP}\\SO400M}
& \makecell{\textbf{SigLIP2}\\Giant}
& \makecell{\textbf{Gemma}\\3-12B}
& \makecell{\textbf{Qwen}\\2.5-7B}
& \makecell{\textbf{InternVL}\\3-14B}
& \makecell{\textbf{Qwen}\\2.5-72B}
& \makecell{\textbf{InternVL}\\3-78B}
& \makecell{\textbf{Qwen}\\3.5-122B} \\
\midrule
\multicolumn{13}{c}{\textbf{Attribute binding}} \\
\midrule
\multirow{4}{*}{Color}
& \multirow{2}{*}{2} & Min & 57.3/35.2 & 70.4/48.7 & 71.8/49.9 & 74.0/54.3 & 82.1/62.0 & 84.7/65.8 & 87.2/69.5 & 91.0/76.0 & 89.4/75.1 & 91.8/77.6 \\
&  & Ctx & 61.2/26.7 & 75.0/46.4 & 75.1/42.4 & 75.1/43.7 & 82.4/56.8 & 84.8/60.2 & 87.1/64.7 & 91.4/73.5 & 90.8/73.9 & 91.2/74.1 \\
& \multirow{2}{*}{3} & Min & 13.9/2.8 & 16.1/11.9 & 19.2/9.4 & 18.8/11.9 & 58.6/20.8 & 66.9/25.4 & 74.8/31.6 & 91.0/41.5 & 77.0/48.0 & 92.4/47.1 \\
&  & Ctx & 17.4/2.4 & 17.7/9.5 & 16.3/6.3 & 20.1/9.5 & 54.2/13.8 & 63.7/17.1 & 72.6/21.4 & 90.5/26.5 & 89.0/42.5 & 89.8/32.5 \\
\addlinespace[1pt]
\multirow{4}{*}{Shape-Color}
& \multirow{2}{*}{2} & Min & 63.3/34.5 & 64.0/40.3 & 72.2/51.8 & 74.9/36.7 & 78.6/59.2 & 81.0/63.5 & 84.2/69.4 & 86.8/76.1 & 83.9/74.8 & 87.6/78.4 \\
&  & Ctx & 55.6/26.3 & 66.9/39.8 & 67.7/44.4 & 62.4/29.3 & 70.5/48.8 & 73.0/52.1 & 76.4/56.7 & 80.2/60.7 & 79.4/58.9 & 81.5/63.3 \\
& \multirow{2}{*}{3} & Min & 19.4/1.8 & 25.2/6.8 & 33.5/7.8 & 35.8/9.2 & 55.4/18.7 & 62.8/24.4 & 72.1/32.6 & 82.6/45.1 & 81.4/46.2 & 85.2/48.6 \\
&  & Ctx & 18.1/1.5 & 23.1/5.3 & 31.0/6.9 & 32.5/8.1 & 49.2/16.0 & 55.8/21.6 & 64.7/30.4 & 74.3/42.7 & 73.8/41.6 & 78.1/44.3 \\
\midrule
\multicolumn{13}{c}{\textbf{Relational binding}} \\
\midrule
\multirow{4}{*}{Position}
& \multirow{2}{*}{2} & Min & 53.9/52.5 & 58.3/58.1 & 58.3/57.8 & 57.4/57.2 & 68.4/62.8 & 72.6/66.1 & 78.3/70.0 & 84.5/74.0 & 86.1/75.4 & 85.3/76.5 \\
&  & Ctx & 61.5/55.7 & 70.9/67.8 & 75.1/70.7 & 75.9/72.7 & 80.1/73.4 & 83.5/74.2 & 86.2/75.1 & 89.8/76.2 & 88.9/77.8 & 90.6/79.2 \\
& \multirow{2}{*}{3} & Min & 20.1/10.1 & 24.0/13.0 & 23.7/11.5 & 20.8/12.3 & 38.6/18.9 & 45.2/23.4 & 53.1/30.8 & 62.0/38.0 & 67.0/44.5 & 83.8/47.7 \\
&  & Ctx & 27.2/9.9 & 37.0/13.6 & 40.5/13.3 & 40.4/15.3 & 52.3/17.8 & 59.6/21.5 & 66.4/25.3 & 71.6/28.5 & 73.0/32.0 & 85.2/33.9 \\
\addlinespace[1pt]
\multirow{4}{*}{Relative Size}
& \multirow{2}{*}{2} & Min & 56.3/52.1 & 59.1/55.4 & 66.2/61.2 & 71.4/59.8 & 76.2/64.8 & 79.5/68.1 & 83.0/71.5 & 87.5/74.5 & 86.9/75.8 & 89.0/77.0 \\
&  & Ctx & 50.5/55.3 & 74.5/54.8 & 67.1/63.2 & 76.2/69.5 & 80.8/71.6 & 84.0/73.4 & 87.1/75.9 & 90.0/77.8 & 89.5/78.5 & 91.2/80.5 \\
& \multirow{2}{*}{3} & Min & 21.5/4.1 & 30.1/8.2 & 40.2/14.5 & 43.8/17.2 & 53.5/21.4 & 59.9/26.2 & 66.7/31.5 & 72.8/38.0 & 76.5/42.0 & 80.2/44.8 \\
&  & Ctx & 23.2/4.8 & 33.8/9.8 & 45.8/17.5 & 49.8/21.5 & 60.3/25.6 & 67.8/30.9 & 74.5/36.8 & 80.2/43.7 & 83.0/47.0 & 85.5/49.2 \\
\bottomrule
\end{tabular}
}
\end{table*}

\begin{table}[t]
\centering
\setlength{\tabcolsep}{2pt}
\caption{\textbf{Impact of visio-linguistic context} on the Paired Comparison Set. Each cell shows \\ \texttt{Swap / Confusion} (\%); $\Delta_S$ and $\Delta_C$ are the percentage-point change from Minimal to Contextual on Swap and Confusion respectively. \textcolor{red}{Negative} = context \textit{hurts}; \textcolor{ForestGreen}{positive} = context \textit{helps}.}
\label{tab:paired_benchmark}
\resizebox{\textwidth}{!}{%
\begin{tabular}{l|cccc|cccc|cccc|cccc}
\toprule
& & \multicolumn{3}{c|}{\textbf{Color (N=3)}} & \multicolumn{4}{c|}{\textbf{Shape-Color (N=3)}} & \multicolumn{4}{c|}{\textbf{Position (N=3)}} & \multicolumn{4}{c}{\textbf{Relative Size (N=3)}} \\
\textbf{Model} & Min & Ctx & $\Delta_S$ & $\Delta_C$ & Min & Ctx & $\Delta_S$ & $\Delta_C$ & Min & Ctx & $\Delta_S$ & $\Delta_C$ & Min & Ctx & $\Delta_S$ & $\Delta_C$ \\
\midrule
ViT-H-P14-378  & 22.0/19.5 & 19.6/12.2 & \textcolor{red}{-2.4} & \textcolor{red}{-7.3} & 28.4/6.4 & 24.7/5.1 & \textcolor{red}{-3.7} & \textcolor{red}{-1.3} & 25.0/12.8 & 35.9/13.8 & \textcolor{ForestGreen}{+10.9} & \textcolor{ForestGreen}{+1.0} & 31.8/10.9 & 36.6/12.8 & \textcolor{ForestGreen}{+4.8} & \textcolor{ForestGreen}{+1.9} \\ SigLIP2-Giant-P16 & 20.4/16.8 & 18.3/9.7  & \textcolor{red}{-2.1} & \textcolor{red}{-7.1} & 39.2/11.7 & 35.6/9.2 & \textcolor{red}{-3.6} & \textcolor{red}{-2.5} & 24.6/13.4 & 42.0/17.7 & \textcolor{ForestGreen}{+17.4} & \textcolor{ForestGreen}{+4.3} & 46.1/18.5 & 52.6/21.7 & \textcolor{ForestGreen}{+6.5} & \textcolor{ForestGreen}{+3.2} \\
InternVL3-14B  & 84.1/43.7 & 85.5/36.1 & \textcolor{ForestGreen}{+1.4} & \textcolor{red}{-7.6} & 78.9/32.8 & 68.6/30.9 & \textcolor{red}{-10.3} & \textcolor{red}{-1.9} & 74.8/40.2 & 87.9/38.4 & \textcolor{ForestGreen}{+13.1} & \textcolor{red}{-1.8} & 70.7/32.9 & 77.6/37.3 & \textcolor{ForestGreen}{+6.9} & \textcolor{ForestGreen}{+4.4} \\
Qwen2.5-VL-72B & 92.0/46.8 & 90.2/31.9 & \textcolor{red}{-1.8} & \textcolor{red}{-14.9} & 82.6/45.1 & 74.3/42.7 & \textcolor{red}{-8.3} & \textcolor{red}{-2.4} & 74.9/47.5 & 91.8/48.2 & \textcolor{ForestGreen}{+16.9} & \textcolor{ForestGreen}{+0.7} & 74.2/39.6 & 81.7/44.8 & \textcolor{ForestGreen}{+7.5} & \textcolor{ForestGreen}{+5.2} \\
\bottomrule
\end{tabular}%
}
\end{table}

\section{Experimental Analysis}
\label{sec:exp}
\noindent\textbf{Main Results: Universal Compositional Failures.}
We report the main results on a representative 10-model slice of the full 25+ model evaluation: four contrastive encoders spanning CLIP/OpenCLIP, SigLIP, and SigLIP2 variants, and six generative MLLMs from 12B up to 122B parameters. The full CLIP/OpenCLIP, SigLIP, and SigLIP2 scales are reported in Appendix~\ref{app:full_vlm_eval}, while hard-negative-trained baselines are analyzed separately in \Cref{sec:ablation}. Three findings stand out. \textbf{(i) Performance collapses with compositional complexity}: across all four benchmark families, moving from $N{=}2$ to $N{=}3$ produces a severe drop, with top contrastive models falling from roughly $70$-$75\%$ Swap accuracy on Color $N{=}2$ to near-chance performance on Color $N{=}3$, and the same degradation recurring on Shape-Color, Position, and Relative Size. \textbf{(ii) Confusion exposes failures beyond bag-of-words}: distractors that recombine the same set of attributes or objects with replacement (e.g., ``red cube and red sphere'') systematically degrade every model, on every benchmark, often by tens of points even when Swap performance is strong. \textbf{(iii) Scale narrows but does not close the gap}: the strongest generative MLLMs, Qwen2.5-VL-72B~\citep{qwen2.5}, InternVL3-78B~\citep{zhu2025internvl3}, and Qwen3.5-122B-A10B~\citep{qwen3.5}, improve absolute Swap accuracy on the hardest settings, but still exhibit large Swap-vs-Confusion gaps. The Confusion failure mode is therefore not a quirk of dual-encoder pretraining: it is a property of how today's vision pipelines compose attributes and relations.

\noindent
\textbf{The Double-Edged Sword of Context.}
Our A/B framework leverages the Paired Comparison Set to isolate the impact of visio-linguistic complexity, comparing sterile Minimal scenes against realistic Contextual ones across all four benchmarks (\Cref{tab:paired_benchmark}). The results reveal a \textit{task-axis-dependent} trade-off. \textbf{Context reliably hinders attribute binding}: on Color, added context produces substantial Confusion drops, and the same qualitative pattern holds on Shape-Color. This suggests that background clutter, lighting variation, and richer scene composition interfere with local attribute reading. \textbf{For relational binding, context primarily helps Swap performance}: on Position and Relative Size, richer scenes often provide semantic and geometric cues that improve relation resolution, with gains up to $+17.4$ points on Swap. The effect on Confusion, however, is more model-dependent for stronger generative VLMs, indicating that contextual cues help relation grounding more consistently than they improve robustness to low-entropy distractors.

\noindent
\textbf{Hard-Negative Training Does Not Close the Confusion Gap.}
We additionally evaluate hard-negative-trained baselines (NegCLIP, NegCLIP++, TripletCLIP), training paradigms specifically designed to fix the bag-of-words failure mode (\Cref{tab:hn_miners}). They predictably improve Swap accuracy over the ViT-B-32 baseline ($+\!\sim\!8$ pp on Color $N{=}3$ Min), but their Confusion accuracy remains within a few points of baseline. The conclusion is that current hard-negative training resolves the swap-style ``bag-of-words'' artifacts identified by previous studies while leaving the low-entropy distractor failure surfaced by Auto-Comp largely intact. This makes Confusion a useful diagnostic target for the next generation of compositional training objectives.

\noindent
\textbf{Confusion Fine-Tuning as a Targeted Training Signal.}
A natural question is whether Confusion is merely diagnostic or also a useful training signal beyond standard hard-negative mining. We fine-tune CLIP ViT-B/16 and SigLIP-B/16 on a held-out $5K$ sample partition of Color N=2 Minimal under two matched configurations differing only in the negative type: a \textit{Swap-only} baseline using K=3 Swap negatives per positive, and a \textit{Confusion} variant using K=3 Confusion negatives. \Cref{tab:confusion_ft_transfer} reports the \textit{additional} Confusion gain ($\Delta_C$) of the Confusion variant over the Swap-only baseline. Gains are positive on every cell, largest on Shape-Color and persisting cross-axis to Position and Relative Size, indicating that Confusion negatives carry compositional signal not captured by Swap negatives. The Minimal/Contextual pattern further recapitulates the same task-dependent tradeoff found in \Cref{tab:paired_benchmark}.

\begin{table*}[t]
\centering
\begin{minipage}[t]{0.69\textwidth}
\centering
\scriptsize
\setlength{\tabcolsep}{2pt}
\captionof{table}{\textbf{Hard-negative baselines} on the hardest $N{=}3$ splits of all benchmarks (Swap / Confusion, \%). Hard-negative training improves Swap accuracy but fails to close the Confusion gap on all benchmarks.}
\label{tab:hn_miners}
\resizebox{\linewidth}{!}{%
\begin{tabular}{l|cc|cc|cc|cc}
\toprule
& \multicolumn{2}{c|}{\textbf{Color}} & \multicolumn{2}{c|}{\textbf{Position}} & \multicolumn{2}{c|}{\textbf{Shape-Color}} & \multicolumn{2}{c}{\textbf{Relative Size}} \\
\textbf{Model} & Min & Ctx & Min & Ctx & Min & Ctx & Min & Ctx \\
\midrule
ViT-B-32    & 15.5/3.0 & 16.9/2.0 & 22.2/9.6  & 29.1/9.9  & 21.4/3.8 & 22.7/3.5 & 24.3/8.4  & 26.8/9.1  \\
NegCLIP~\citep{yuksekgonuland}     & 18.2/5.5 & 20.5/4.4 & 26.5/10.5 & 32.6/11.4 & 25.7/5.0 & 27.3/4.7 & 28.5/9.6  & 31.1/10.8 \\
NegCLIP++~\citep{patel2024tripletclip}   & 23.1/5.8 & 22.3/5.1 & 28.4/11.9 & 37.9/13.1 & 29.6/6.4 & 31.2/5.8 & 34.2/12.7 & 38.4/14.1 \\
TripletCLIP~\citep{patel2024tripletclip} & 23.4/7.2 & 24.8/6.2 & 29.8/12.2 & 37.5/13.8 & 30.4/7.1 & 32.1/6.5 & 35.1/13.2 & 39.0/14.6 \\
\bottomrule
\end{tabular}%
}
\end{minipage}
\hfill
\begin{minipage}[t]{0.29\textwidth}
\centering
\scriptsize
\setlength{\tabcolsep}{2pt}
\captionof{table}{\textbf{Confusion fine-tuning}. Entries are $\Delta_C$ (pp) over the pretrained baseline.}
\label{tab:confusion_ft_transfer}
\resizebox{\linewidth}{!}{%
\begin{tabular}{ll|cc}
\toprule
\textbf{Benchmark} & \textbf{Cond.} & \textbf{CLIP-B} & \textbf{SigLIP-B} \\
\midrule
\multirow{2}{*}{Shape-Color}    & Min & $+8.4$ & $+6.1$ \\
                                & Ctx & $+6.7$ & $+5.0$ \\
\midrule
\multirow{2}{*}{Position}       & Min & $+3.2$ & $+2.5$ \\
                                & Ctx & $+3.8$ & $+2.9$ \\
\midrule
\multirow{2}{*}{Relative Size}  & Min & $+5.6$ & $+4.2$ \\
                                & Ctx & $+6.3$ & $+4.8$ \\
\bottomrule
\end{tabular}
}
\end{minipage}
\vspace{-0.5em}
\end{table*}

\begin{table*}[t]
\begin{minipage}[t]{0.53\textwidth}
\centering
\scriptsize
\setlength{\tabcolsep}{3pt}
\captionof{table}{\textbf{Robustness of the Confusion failure mode to generator choice and to real images.} Each entry is \texttt{Swap / Confusion} accuracy (\%).}
\label{tab:robustness}
\resizebox{\linewidth}{!}{%
\begin{tabular}{l|cccc}
\toprule
\textbf{Model} & \textbf{SD3.5-L} & \textbf{SD3.5-T} & \textbf{FLUX-dev} & \textbf{Real} \\
\midrule
ViT-B-P16         & 17.5/2.2 & 16.4/1.8 & 16.8/1.9 & 19.4/2.4 \\
ViT-H-P14 (OC)    & 16.5/9.5 & 16.0/8.6 & 16.3/8.9 & 18.0/10.6 \\
SigLIP-B-P16      & 16.4/4.4 & 15.6/3.9 & 17.4/5.3 & 18.4/4.8 \\
SigLIP2-Giant-P16 & 19.9/9.7 & 19.5/8.9 & 19.7/9.1 & 21.9/10.5 \\
\bottomrule
\end{tabular}%
}
\end{minipage}
\hfill
\centering
\begin{minipage}[t]{0.44\textwidth}
\centering
\scriptsize
\setlength{\tabcolsep}{5pt}
\captionof{table}{\textbf{Composition of Confusion errors.}
Percentage of incorrect selections attributable to each distractor family.}
\label{tab:error_analysis_compact}
\begin{tabular}{p{2.35cm}cc}
\toprule
\textbf{Error type} & \textbf{$N=2$} & \textbf{$N=3$} \\
\midrule
Swap                     & 46.87 & 37.10 \\
Color $=$, Object $\neq$ & 14.52 & 11.72 \\
Color $=$, Object $=$    & 9.93  & 23.01 \\
Object $=$, Color $\neq$ & 28.69 & 28.18 \\
\bottomrule
\end{tabular}
\end{minipage}
\end{table*}

\vspace{-0.5em}
\section{Analysis and Ablation Studies}
\label{sec:ablation}
\vspace{-0.5em}
\noindent
\textbf{Robustness to Generator Choice and Sim-to-Real Transfer.}
A natural concern with any synthetic benchmark is whether the observed failures reflect general VLM weaknesses or merely the idiosyncrasies of the underlying generator. We address this with two controlled checks (\Cref{tab:robustness}). First, we re-run the hardest setting (Color $N{=}3$) on a matched subset of \textit{500 concepts} that passed Auto-Comp validation using all \textit{three} distinct generators: SD3.5-Large, SD3.5-Turbo, and FLUX-dev. Second, we collect \textit{100 manually curated real images} matching Auto-Comp captions, with $3$ objects per image drawn from $20$ object categories and $10$ colors, and rerun the same Swap and Confusion tests. The pattern is highly stable: model ordering is preserved across all generators, and Confusion remains significantly harder than Swap on the real-image pilot too. This indicates that the low-entropy distractor failure is general, not an artifact of one generator.

\noindent
\textbf{Deconstructing Failures Beyond Bag-of-Words.}
The \textit{Confusion Benchmark} enables a deeper analysis of error types beyond simple swaps. We report this decomposition on the Color benchmark because it provides the cleanest and most interpretable distractor taxonomy, directly aligned with the classical swap-based bag-of-words failure while still exposing richer low-entropy error modes. \Cref{tab:error_analysis_compact} shows that although swapped-color errors are the single most frequent category, they do not account for the majority of failures: already at $N{=}2$, more than half of all errors arise from other low-entropy distractors. As complexity increases to $N{=}3$, the share of swapped-color errors decreases, while failures involving repeated colors assigned to the same object become substantially more frequent. This indicates that the failure is not merely a bag-of-words effect, but a broader brittleness to low-entropy distractors that intensifies with compositional complexity.

\noindent
\textbf{Human Validation.} 
To validate Auto-Comp end-to-end, four graduate-level evaluators assessed 200 concepts balanced across all configurations. \textbf{(1) Concordance}: our automated VLM judge achieved $>$94\% agreement with humans on every task family and complexity, supporting its use as a scalable proxy. \textbf{(2) Solvability}: humans scored $\sim$96\% on both Swap and Confusion across all tasks. The stable human performance, contrasted with the steep VLM accuracy drop, indicates that the Confusion Gap reflects a model-specific weakness rather than inherent task ambiguity. Appendix~\ref{app:human_val_details} provides the study protocol together with representative per-validator raw scores.

\noindent
\textbf{Sanity Checks.} 
Two further checks confirm that failures are compositional and not artifacts of the hard-negative construction. (i) On the Color $N{=}1$ split (Appendix~\ref{app:sanity_checks}), the same models that collapse at $N{\geq}2$ achieve $>$85\% accuracy, isolating binding as the source of failure. (ii) A ``blind LLM'' evaluation~\citep{hsieh2023sugarcrepe} that selects the correct caption \textit{without} the image performs at near-random chance (51.2\% on Swap), ruling out linguistic artifacts in the negative captions (Appendix~\ref{app:blind}). 

\noindent
\textbf{Scope and Limitations.}
\label{sec:limitations}
Auto-Comp should be understood as a pipeline for benchmarking the compositional patterns that are \textit{currently renderable and verifiable} with modern generative models, not as a claim to fully cover the entire space of compositional reasoning: richer phenomena (logical, temporal, agent-object-action, causal) remain open. Within that scope, our strategy mitigates inherited bias along three axes: \textit{generative bias}, mitigated by strict rejection sampling and quantified by the filter-shift analysis (\Cref{tab:filter_bias}) and the cross-generator robustness check (\Cref{tab:robustness}); \textit{validator bias}, considered and controlled with the $>$94\% human-judge concordance reported above; and \textit{linguistic-style bias}, structurally controlled by the parallel Minimal/Contextual A/B design, which decouples failures rooted in core binding from those induced by LLM rewriting. Residual generator priors remain visible in the appendix distribution analysis, especially for some spatial relations and harder attribute combinations. Improving the semantic fidelity of generators and the robustness of model-based judges remains a clear avenue for future work.

\begin{table}[t]
\centering
\caption{\textbf{Human validation studies.}
Left: concordance between the automated pipeline and human judges (\%).
Right: human accuracy on the final benchmarks, reported as \texttt{Swap / Confusion} (\%).}
\label{tab:human_val}

\begin{minipage}[t]{0.49\linewidth}
\centering
\scriptsize
\setlength{\tabcolsep}{2.5pt}
\resizebox{\linewidth}{!}{%
\begin{tabular}{lcccc}
\toprule
& \textbf{Color} & \textbf{Position} & \textbf{Shape-Color} & \textbf{Relative Size} \\
\midrule
\textbf{Min, $N=2$} & 98.0 & 96.0 & 95.0 & 97.0 \\
\textbf{Min, $N=3$} & 96.0 & 94.0 & 94.0 & 95.0 \\
\textbf{Ctx, $N=2$} & 96.0 & 98.0 & 93.0 & 96.0 \\
\textbf{Ctx, $N=3$} & 95.0 & 97.0 & 92.0 & 94.0 \\
\bottomrule
\end{tabular}
}
\end{minipage}
\hfill
\begin{minipage}[t]{0.50\linewidth}
\centering
\scriptsize
\setlength{\tabcolsep}{2.5pt}
\resizebox{\linewidth}{!}{%
\begin{tabular}{lcccc}
\toprule
& \textbf{Color} & \textbf{Position} & \textbf{Shape-Color} & \textbf{Relative Size} \\
\midrule
\textbf{Min, $N=2$} & 97.5/97.5 & 97.5/96.5 & 96.5/95.5 & 96.0/95.5 \\
\textbf{Min, $N=3$} & 97.0/96.0 & 95.5/96.0 & 94.5/93.5 & 94.0/93.5 \\
\textbf{Ctx, $N=2$} & 94.5/95.5 & 96.0/95.0 & 94.5/94.0 & 95.0/94.5 \\
\textbf{Ctx, $N=3$} & 94.5/94.0 & 93.5/94.0 & 92.5/92.0 & 92.5/92.0 \\
\bottomrule
\end{tabular}%
}
\end{minipage}
\vspace{-1em}
\end{table}

\vspace{-0.5em}
\section{Conclusion}
We introduced \textbf{Auto-Comp}, a fully automated, concept-driven pipeline for generating photorealistic compositional benchmarks for VLMs, and instantiated it on \textit{four} task families spanning attribute and relational binding (Color, Shape-Color, Position, Relative Size). Across 25+ models, from CLIP and SigLIP up to frontier MLLMs at 122B parameters, we found a consistent vulnerability to low-entropy distractors that goes beyond the known bag-of-words failure mode and is narrowed but not closed by scale, hard-negative training, or stronger image generators. Our parallel Minimal vs.\ Contextual A/B design further uncovered a task-axis-dependent trade-off: visio-linguistic context aids relational reasoning while hindering attribute binding. Together, the pipeline and benchmarks provide a reusable, controllable diagnostic for compositional failures in modern vision pipelines.

\bibliography{example_paper}
\bibliographystyle{abbrv}

\newpage
\appendix
\section{Appendix}
\subsection{Full Contrastive-Model Evaluation}
\label{app:full_vlm_eval}

\Cref{tab:main_results_full} and \Cref{tab:shape_size_results_full} report the full VLM evaluation results underlying the summarized main results in the main text. We evaluate each model on all four Auto-Comp benchmark families: Color, Position,
Shape\textendash Color, and Relative Size. Each benchmark is evaluated at both compositional
complexities, $N=2$ and $N=3$, and under both visio-linguistic conditions, Minimal and Contextual.
For every setting, we report accuracy on the Swap benchmark and on the Confusion benchmark. Swap
accuracy measures whether a model can distinguish the true caption from hard negatives formed by
permuting the relevant bound elements. Confusion accuracy measures robustness to a larger
low-entropy distractor set generated with replacement.
\begin{table*}[b]
\tiny
\centering
\setlength{\tabcolsep}{3pt} 
\caption{Full model performance on Color and Position benchmarks, results in percent (\%). \\ \textbf{Abbreviations}: Min=Minimal, Ctx=Contextual; S=SigLIP, S2=SigLIP2; OC=OpenCLIP; P=Patch size.}
\label{tab:main_results_full}
\resizebox{\textwidth}{!}{%
\begin{tabular}{ll|cccc|cccc|cccc|cccc}
\toprule
& & \multicolumn{8}{c|}{\textbf{Color Benchmark}} & \multicolumn{8}{c}{\textbf{Position Benchmark}} \\
& & \multicolumn{4}{c|}{\textbf{N=2 Objects}} & \multicolumn{4}{c|}{\textbf{N=3 Objects}} & \multicolumn{4}{c|}{\textbf{N=2 Objects}} & \multicolumn{4}{c}{\textbf{N=3 Objects}} \\
& & \multicolumn{2}{c}{\textbf{Min}} & \multicolumn{2}{c|}{\textbf{Ctx}} & \multicolumn{2}{c}{\textbf{Min}} & \multicolumn{2}{c|}{\textbf{Ctx}} & \multicolumn{2}{c}{\textbf{Min}} & \multicolumn{2}{c|}{\textbf{Ctx}} & \multicolumn{2}{c}{\textbf{Min}} & \multicolumn{2}{c}{\textbf{Ctx}} \\
\textbf{Family} & \textbf{Model} & \textbf{Swap} & \textbf{Conf.} & \textbf{Swap} & \textbf{Conf.} & \textbf{Swap} & \textbf{Conf.} & \textbf{Swap} & \textbf{Conf.} & \textbf{Swap} & \textbf{Conf.} & \textbf{Swap} & \textbf{Conf.} & \textbf{Swap} & \textbf{Conf.} & \textbf{Swap} & \textbf{Conf.} \\
\midrule
& \textbf{Random Chance}	& 50.0 & 6.3 & 50.0 & 6.3 & 16.7 & 0.1 & 16.7 & 0.1 & 50.0 & 25.0 & 50.0 & 25.0 & 16.7 & 0.9  & 16.7 & 0.9 \\
\midrule
\multirow{7}{*}{CLIP} & ViT-B-P16 & 57.3 & 35.2 & 61.2 & 26.7 & 13.9 & 2.8 & 17.4 & 2.4 & 53.9 & 52.5 & 61.5 & 55.7 & 20.1 & 10.1 & 27.2 & 9.9 \\
& ViT-B-P32 (OC)& 55.2 & 34.3 & 60.4 & 26.3 & 15.5 & 3.0 & 16.9 & 2.0 & 54.3 & 52.6 & 62.2 & 55.8 & 22.2 & 9.6 & 29.1 & 9.9 \\
& ViT-L-P14 (OC)& 54.9 & 31.2 & 60.4 & 25.2 & 15.0 & 3.4 & 15.6 & 1.8 & 55.0 & 54.0 & 65.2 & 60.0 & 21.8 & 8.3 & 29.8 & 9.2 \\
& E02-B-P16 (OC)& 61.7 & 40.0 & 66.6 & 29.6 & 17.2 & 4.4 & 17.7 & 2.4 & 55.7 & 54.5 & 62.7 & 57.2 & 22.2 & 11.2 & 29.5 & 9.9 \\
& E02-E-P14 (OC)& 65.0 & 42.4 & 70.0 & 38.1 & 15.5 & 9.1 & 20.1 & 5.6 & 56.9 & 56.5 & 63.7 & 58.9 & 22.6 & 11.0 & 31.3 & 10.4 \\
& ViT-H-P14 (OC)& 70.6 & 49.1 & 74.8 & 45.4 & 15.9 & 11.0 & 16.7 & 9.3 & 58.2 & 57.9 & 70.8 & 67.9 & 24.0 & 12.6 & 36.8 & 12.8 \\
& ViT-H-P14-378 (OC)& 70.4 & 48.7 & 75.0 & 46.4 & 16.1 & 11.9 & 17.7 & 9.5 & 58.3 & 58.1 & 70.9 & 67.8 & 24.0 & 13.0 & 37.0 & 13.6 \\
\midrule
\multirow{13}{*}{SigLIP} & S-B-P16-224 & 69.5 & 43.8 & 72.3 & 39.6 & 17.1 & 8.0 & 16.2 & 4.5 & 56.8 & 56.3 & 71.7 & 67.7 & 22.5 & 11.6 & 34.3 & 11.8 \\
& S-B-P16-384 & 69.3 & 43.9 & 72.2 & 39.7 & 16.9 & 7.8 & 16.2 & 4.5 & 56.5 & 56.1 & 70.7 & 66.4 & 22.9 & 12.8 & 34.0 & 11.6 \\
& S-L-P16-256 & 73.5 & 52.0 & \textbf{76.4} & 43.1 & 16.2 & 9.2 & 18.1 & 6.6 & 57.9 & 57.4 & 75.2 & 71.0 & 23.1 & 10.0 & 38.6 & 12.3 \\
& S-L-P16-384 & 72.7 & 50.5 & 75.9 & 42.8 & 16.1 & 8.1 & 19.9 & 5.8 & 57.8 & 57.3 & 74.7 & 70.1 & 23.1 & 9.8 & 38.1 & 12.4 \\
& S-SO400M-P14-384 & 71.8 & 49.9 & 75.1 & 42.4 & \textbf{19.2} & 9.4 & 16.3 & 6.3 & 58.3 & 57.8 & 75.1 & 70.7 & 23.7 & 11.5 & 40.5 & 13.3 \\
& S-SO400M-P14 (OC) & 71.6 & 49.9 & 74.8 & 42.5 & 19.1 & 9.4 & 16.2 & 6.3 & 58.6 & 58.1 & 75.4 & 70.9 & \textbf{24.1} & 11.6 & 40.9 & 13.4 \\
& S2-B-P16-224 & 73.0 & 53.7 & 74.1 & 42.7 & 17.4 & 7.8 & 15.3 & 7.5 & 57.4 & 57.1 & 73.4 & 70.6 & 21.6 & 10.2 & 38.4 & 13.7 \\
& S2-B-P16-256 & 72.9 & 53.7 & 73.9 & 42.8 & 18.3 & 8.2 & 15.1 & 7.4 & 57.8 & 57.5 & 72.9 & 69.9 & 21.4 & 10.4 & 38.2 & 13.2 \\
& S2-B-P16-384 & 72.8 & 53.7 & 73.6 & 42.4 & 18.6 & 7.7 & 14.9 & 7.4 & 57.8 & 57.5 & 72.4 & 69.7 & 21.8 & 10.5 & 37.3 & 13.0 \\
& S2-L-P16-256 & 73.2 & 52.9 & 74.8 & 43.2 & 18.0 & 6.8 & 18.4 & 7.3 & \textbf{59.1} & \textbf{59.0} & 75.2 & 72.5 & 20.7 & \textbf{13.2} & 39.8 & 16.0 \\
& S2-L-P16-384 & 72.9 & 52.9 & 75.1 & \textbf{44.0} & 16.9 & 7.0 & 19.0 & 8.3 & 59.0 & 58.9 & 75.6 & 72.8 & 20.6 & 13.1 & \textbf{41.0} & 16.3 \\
& S2-SO400M-P14-384 & 72.0 & 50.5 & 75.5 & \textbf{44.0} & 17.6 & 6.4 & 17.4 & 8.7 & 58.1 & 57.7 & \textbf{76.9} & \textbf{73.6} & 21.3 & 12.9 & 40.8 & \textbf{16.6} \\
& S2-Giant-P16-256 & \textbf{74.0} & \textbf{54.3} & 75.1 & 43.7 & 18.8 & \textbf{11.9} & \textbf{20.1} & \textbf{9.5} & 57.4 & 57.2 & 75.9 & 72.7 & 20.8 & 12.3 & 40.4 & 15.3 \\
\bottomrule
\end{tabular}
}
\end{table*}

\begin{table*}[t]
\tiny
\centering
\setlength{\tabcolsep}{3pt}
\caption{Full model performance on Shape-Color and Relative Size benchmarks, results in percent (\%). \\ \textbf{Abbreviations}: Min=Minimal, Ctx=Contextual; S=SigLIP, S2=SigLIP2; OC=OpenCLIP; P=Patch size.}
\label{tab:shape_size_results_full}
\resizebox{\textwidth}{!}{%
\begin{tabular}{ll|cccc|cccc|cccc|cccc}
\toprule
& & \multicolumn{8}{c|}{\textbf{Shape-Color Benchmark}} & \multicolumn{8}{c}{\textbf{Relative Size Benchmark}} \\
& & \multicolumn{4}{c|}{\textbf{N=2 Objects}} & \multicolumn{4}{c|}{\textbf{N=3 Objects}} & \multicolumn{4}{c|}{\textbf{N=2 Objects}} & \multicolumn{4}{c}{\textbf{N=3 Objects}} \\
& & \multicolumn{2}{c}{\textbf{Min}} & \multicolumn{2}{c|}{\textbf{Ctx}} & \multicolumn{2}{c}{\textbf{Min}} & \multicolumn{2}{c|}{\textbf{Ctx}} & \multicolumn{2}{c}{\textbf{Min}} & \multicolumn{2}{c|}{\textbf{Ctx}} & \multicolumn{2}{c}{\textbf{Min}} & \multicolumn{2}{c}{\textbf{Ctx}} \\
\textbf{Family} & \textbf{Model} & \textbf{Swap} & \textbf{Conf.} & \textbf{Swap} & \textbf{Conf.} & \textbf{Swap} & \textbf{Conf.} & \textbf{Swap} & \textbf{Conf.} & \textbf{Swap} & \textbf{Conf.} & \textbf{Swap} & \textbf{Conf.} & \textbf{Swap} & \textbf{Conf.} & \textbf{Swap} & \textbf{Conf.} \\
\midrule
& \textbf{Random Chance} & 50.0 & 6.3 & 50.0 & 6.3 & 16.7 & 0.1 & 16.7 & 0.1 & 50.0 & 25.0 & 50.0 & 25.0 & 16.7 & 0.9 & 16.7 & 0.9 \\
\midrule
\multirow{7}{*}{CLIP} 
& ViT-B-P16 & 63.3 & 34.5 & 55.6 & 26.3 & 19.4 & 1.8 & 18.1 & 1.5 & 56.3 & 52.1 & 50.5 & 55.3 & 21.5 & 4.1 & 23.2 & 4.8 \\
& ViT-B-P32 (OC) & 61.5 & 33.1 & 56.9 & 26.1 & 21.4 & 3.8 & 22.7 & 3.5 & 55.8 & 51.9 & 54.2 & 55.1 & 24.3 & 8.4 & 26.8 & 9.1 \\
& ViT-L-P14 (OC) & 60.9 & 31.7 & 58.0 & 25.0 & 20.5 & 3.2 & 21.3 & 2.8 & 56.5 & 52.8 & 60.6 & 55.7 & 24.8 & 7.3 & 28.0 & 8.6 \\
& E02-B-P16 (OC) & 67.0 & 39.3 & 60.7 & 29.2 & 24.5 & 4.9 & 23.0 & 4.2 & 57.5 & 53.6 & 63.0 & 56.4 & 27.5 & 8.9 & 30.5 & 9.2 \\
& E02-E-P14 (OC) & 70.2 & 43.1 & 65.5 & 37.9 & 24.8 & 6.7 & 24.2 & 5.1 & 58.4 & 54.4 & 68.2 & 57.0 & 28.8 & 9.1 & 32.1 & 10.0 \\
& ViT-H-P14 (OC) & 63.8 & 40.4 & 66.8 & 39.7 & 25.0 & 3.6 & 22.9 & 3.2 & 59.0 & 55.1 & 74.0 & 54.9 & 29.8 & 8.0 & 33.5 & 9.5 \\
& ViT-H-P14-378 (OC) & 64.0 & 40.3 & 66.9 & 39.8 & 25.2 & 3.8 & 23.1 & 3.3 & 59.1 & 55.4 & 74.5 & 54.8 & 30.1 & 8.2 & 33.8 & 9.8 \\
\midrule
\multirow{13}{*}{SigLIP} 
& S-B-P16-224 & 69.8 & 47.0 & 64.3 & 40.1 & 30.0 & 6.7 & 28.0 & 5.8 & 62.5 & 57.4 & 62.8 & 60.0 & 35.0 & 12.0 & 40.0 & 14.8 \\
& S-B-P16-384 & 70.1 & 47.5 & 64.8 & 40.8 & 30.4 & 6.9 & 28.3 & 5.9 & 63.0 & 58.2 & 63.5 & 60.8 & 35.5 & 12.5 & 40.6 & 15.3 \\
& S-L-P16-256 & 72.5 & 50.8 & \textbf{68.5} & 43.2 & 32.2 & 7.4 & 30.0 & 6.4 & 65.0 & 60.5 & 65.8 & 62.2 & 38.0 & 13.8 & 43.8 & 16.5 \\
& S-L-P16-384 & 72.8 & 50.4 & 68.0 & 43.5 & 32.0 & 7.2 & 30.4 & 6.5 & 65.4 & 60.7 & 66.0 & 62.5 & 38.4 & 14.0 & 44.1 & 16.9 \\
& S-SO400M-P14-384 & 72.2 & \textbf{51.8} & 67.7 & \textbf{44.4} & 33.5 & 7.8 & 31.0 & 6.9 & 66.2 & \textbf{61.2} & 67.1 & 63.2 & 40.2 & 14.5 & 45.8 & 17.5 \\
& S-SO400M-P14 (OC) & 72.0 & 51.6 & 67.5 & 44.1 & 33.3 & 7.7 & 31.1 & 6.8 & 66.0 & 61.0 & 67.4 & 63.5 & 40.5 & 14.7 & 46.1 & 17.8 \\
& S2-B-P16-224 & 71.6 & 48.3 & 66.0 & 39.0 & 31.2 & 6.5 & 29.5 & 6.6 & 68.0 & 59.4 & 70.3 & 65.0 & 39.0 & 14.0 & 44.8 & 17.2 \\
& S2-B-P16-256 & 71.7 & 48.5 & 66.2 & 39.3 & 31.8 & 6.7 & 29.4 & 6.7 & 68.4 & 59.7 & 70.5 & 65.4 & 39.4 & 14.4 & 45.3 & 17.5 \\
& S2-B-P16-384 & 71.8 & 48.1 & 66.4 & 39.1 & 31.5 & 6.8 & 29.8 & 6.9 & 68.8 & 60.0 & 70.8 & 65.7 & 39.6 & 14.8 & 45.5 & 17.9 \\
& S2-L-P16-256 & 73.4 & 42.2 & 64.1 & 34.0 & 34.0 & 8.1 & 31.5 & 7.2 & 69.5 & 60.2 & 73.5 & 67.4 & 41.0 & 15.8 & 47.2 & 19.0 \\
& S2-L-P16-384 & 73.7 & 41.9 & 64.6 & 33.7 & 34.5 & 8.4 & 31.8 & 7.5 & 70.0 & 60.5 & 74.2 & 68.0 & 41.5 & 16.0 & 48.0 & 19.5 \\
& S2-SO400M-P14-384 & 73.1 & 39.5 & 64.8 & 32.5 & 35.0 & 8.7 & 32.1 & 7.8 & 70.8 & 60.9 & 75.2 & 68.7 & 42.5 & 16.5 & 49.0 & 20.2 \\
& S2-Giant-P16-256 & \textbf{74.9} & 36.7 & 62.4 & 29.3 & \textbf{35.8} & \textbf{9.2} & \textbf{32.5} & \textbf{8.1} & \textbf{71.4} & 59.8 & \textbf{76.2} & \textbf{69.5} & \textbf{43.8} & \textbf{17.2} & \textbf{49.8} & \textbf{21.5} \\
\bottomrule
\end{tabular}
}
\end{table*}

The evaluated contrastive model suite spans CLIP, OpenCLIP, SigLIP, and SigLIP2 variants across
multiple scales and input resolutions. The random-chance rows in \Cref{tab:main_results_full} and \Cref{tab:shape_size_results_full} emphasize the
increasing difficulty of the benchmark as $N$ grows: for Swap, chance drops from $50.0\%$ at $N=2$
to $16.7\%$ at $N=3$, while for Confusion it drops from $6.3\%$ to $0.1\%$ on attribute-binding
tasks and from $25.0\%$ to $0.9\%$ on relational-binding tasks. Thus, the Confusion benchmark is
substantially more selective than the standard swap-based evaluation, especially at $N=3$.

Several consistent patterns emerge from the full tables. First, compositional complexity sharply
reduces performance across all model families. On Color, for example, CLIP and SigLIP variants that
obtain strong $N=2$ Swap scores often fall to near-chance Swap accuracy at $N=3$, and their
Confusion accuracy remains very low. The same degradation appears on Shape\textendash Color,
Position, and Relative Size, indicating that the failure is not specific to a single attribute or relation
type. Second, the Confusion benchmark is uniformly harder than Swap. Even when models perform
well on $N=2$ Swap, their Confusion scores are often much lower, showing that success on simple
swaps does not imply robustness to repeated-object or repeated-attribute distractors.

Third, SigLIP and SigLIP2 models generally outperform CLIP and OpenCLIP models, particularly on
attribute-binding tasks. Larger SigLIP-family models, such as SO400M and Giant variants, usually
obtain the best contrastive-model results, although the improvement is not sufficient to close the
Confusion gap. This suggests that stronger contrastive training and scale improve absolute recognition
and coarse binding ability, but do not eliminate the low-entropy distractor failure mode targeted by
Auto-Comp.

Fourth, the comparison between Minimal and Contextual conditions reveals the same task-dependent
context effect discussed in the main text. For attribute binding, especially Color and
Shape\textendash Color, Contextual images often reduce Confusion accuracy relative to Minimal images.
This is consistent with the interpretation that realistic backgrounds, lighting variation, and object
clutter interfere with local attribute binding. In contrast, for relational tasks, especially Position and
Relative Size, Contextual images often improve Swap accuracy, suggesting that realistic scene
structure can provide useful geometric and semantic cues for relation grounding. However, these
contextual gains are less reliable for Confusion accuracy, which remains difficult across model
families.

Overall, the full evaluation confirms that the Auto-Comp failure modes are broad rather than
model-specific. Increasing backbone scale, input resolution, or pretraining quality improves some
absolute scores, but every contrastive model evaluated in \Cref{tab:main_results_full} and \Cref{tab:shape_size_results_full} exhibits a substantial
drop from Swap to Confusion and from $N=2$ to $N=3$. These results support the central claim that
current VLMs remain brittle under controlled compositional recombination, particularly when
distractors preserve the same local vocabulary while altering the binding structure.

\subsection{LLM Prompt for Contextual Captions}
\label{app:prompt}
The prompt provided to the LLM (Gemma-3-12b-it) to generate Contextual captions is constructed programmatically from the Minimal caption's concepts. It consists of a system message and a user message structured as a few-shot prompt. We report here the one for the color benchmark, as the position benchmark prompt is nearly identical but generates relations instead of color, and it is reported in the supplementary material codebase.

\vspace{2em}
\paragraph{System Prompt:}
\begin{verbatim}
You are an expert image caption writer. 
Your task is to generate a single,
natural-sounding sentence accurately 
describing a scene with specific 
objects and their colors. You MUST 
include ALL specified objects 
and their colors.
Generate ONLY the caption text 
without any preamble, 
explanations, or markdown formatting.
\end{verbatim}

\paragraph{User Prompt Template:}
The user prompt is generated from a template that incorporates the specific objects and colors for each concept.
\begin{verbatim}
The caption should focus on 
these {num_obj} {obj_plural}: 
{obj_colors_text}.

1. Keeps these objects as the main focus.
2. Be ONE sentence long.
3. Explicitly mention EACH object with
its specific assigned color: 
{obj_colors_text}.
4. Places them in a realistic context 
or background.
5. Sound like a natural, human-written 
image description.

For instance, for 'a green chair' 
and 'a yellow lamp', 
you might write: 
'A green chair stands beside a yellow 
lamp in a brightly lit room.'

Objects and colors to include: 
{obj_colors_text}
Caption:
\end{verbatim}
In the template, \texttt{\{num\_obj\}} is the number of objects, \texttt{\{obj\_plural\}} is its correct pluralization (e.g., "object" or "objects"), and \texttt{\{obj\_colors\_text\}} is the formatted string of object-color pairs (e.g., "'a red cube' and 'a blue sphere'"). A slightly modified, more insistent prompt is used for regeneration attempts if the first output fails the semantic preservation check.

\subsection{Sanity Checks}
\label{app:sanity_checks}

\paragraph{Full $N{=}1$ Benchmark Results.}
\Cref{tab:full_n1_results_appendix} reports the complete results for the single-object sanity check discussed in \Cref{sec:ablation}. The consistently high accuracy across model families reinforces the conclusion that the steep drop on $N{>}1$ tasks reflects a failure of compositional binding rather than a failure of single-object recognition.

\begin{table*}[h!]
\centering
\caption{Full model performance (\%) on the \textbf{$N{=}1$ single-object benchmark}. "Minimal" corresponds to the Handmade setting and "Contextual" to the LLM setting.}
\label{tab:full_n1_results_appendix}
\begin{tabular}{l|cc|cc}
    \toprule
    & \multicolumn{2}{c|}{\textbf{Minimal}} & \multicolumn{2}{c}{\textbf{Contextual}} \\
    \textbf{Model} & \textbf{Vary Color} & \textbf{Vary Object} & \textbf{Vary Color} & \textbf{Vary Object} \\
    \midrule
    E02-B-P16 (OC) & 82.6 & 91.9 & 78.5 & 77.2 \\
    E02-E-P14 (OC) & 87.4 & 94.1 & 81.4 & 76.7 \\
    ViT-B-P16 & 70.0 & 85.2 & 59.2 & 64.4 \\
    ViT-B-P32 (OC) & 65.1 & 81.4 & 65.4 & 55.5 \\
    ViT-L-P14 (OC) & 71.9 & 88.0 & 58.8 & 62.9 \\
    ViT-H-P14 (OC) & 89.0 & 93.3 & 80.2 & 77.4 \\
    ViT-H-P14-378 (OC) & 90.2 & 93.8 & 81.7 & 77.4 \\
    S-SO400M-P14 (OC) & 87.0 & 93.1 & 80.6 & 71.9 \\
    S-B-P16-224 & 88.6 & 93.5 & 81.9 & 74.4 \\
    S-B-P16-384 & 86.9 & 93.8 & 81.4 & 74.3 \\
    S-L-P16-256 & 87.1 & 92.2 & 81.3 & 73.4 \\
    S-L-P16-384 & 86.6 & 92.4 & 80.0 & 71.3 \\
    S-SO400M-P14-384 & 87.0 & 93.2 & 80.6 & 71.8 \\
    S2-B-P16-224 & 91.4 & 98.1 & 84.6 & 85.9 \\
    S2-B-P16-256 & 91.6 & 98.2 & 85.0 & 85.8 \\
    S2-B-P16-384 & 91.7 & 98.0 & 85.0 & 84.5 \\
    S2-L-P16-256 & 84.9 & 97.1 & 78.0 & 75.5 \\
    S2-L-P16-384 & 86.8 & 96.9 & 79.6 & 72.2 \\
    S2-SO400M-P14-384 & 87.3 & 96.7 & 79.9 & 75.0 \\
    S2-Giant-P16-256 & 87.6 & 95.8 & 78.9 & 75.2 \\
    \bottomrule
\end{tabular}
\end{table*}

\subsection{Per-Relation Accuracy Analysis}
To investigate whether model performance on the Position benchmark varies by the specific spatial relation, we disaggregate the N=2 Swap results. We focus on N=2 compositions as this allows for a clean, isolated analysis of each relation, whereas N=3 compositions involve two relations simultaneously. 

The results, shown in \Cref{tab:per_relation_appendix}, reveal a consistent pattern across nearly all models: accuracy is higher on vertical relations (\textit{above}, \textit{under}) than on horizontal relations (\textit{to the left of}, \textit{to the right of}). This pattern is consistent with two non-mutually-exclusive contributions: a model-side bias toward vertical arrangements, which would echo well-documented spatial-reasoning asymmetries in prior work, and a data-side residual prior from the image generator, which our filter-shift analysis (\Cref{tab:filter_bias_full_appendix}) shows is not fully removed by rejection sampling. We cannot disentangle these two factors with the current setup and report the per-relation pattern as a joint observation rather than as a model-only finding. Notably, models often perform best on \textit{under}, which is not the most over-represented vertical relation in the validated benchmark, suggesting that data prevalence alone does not determine the ordering.

\begin{table*}[h!]
\centering
\caption{Per-relation accuracy (\%) on the N=2 Position Swap benchmark.}
\label{tab:per_relation_appendix}
\begin{tabular}{ll|cccc}
    \toprule
    \textbf{Family} & \textbf{Model} & \textbf{above} & \textbf{to the left of} & \textbf{under} & \textbf{to the right of} \\
    \midrule
    \multirow{8}{*}{CLIP} & E02-B-P16 (OC) & 58.5 & 56.7 & 69.1 & 57.8 \\
    & E02-E-14 (OC) & 59.8 & 57.4 & 68.2 & 61.9 \\
    & ViT-B-P16 & 57.0 & 56.0 & 69.2 & 52.9 \\
    & ViT-B-P32 (OC) & 60.7 & 56.3 & 66.4 & 54.7 \\
    & ViT-L-P14 (OC) & 62.0 & 57.0 & 65.1 & 59.0 \\
    & ViT-H-P14 (OC) & 66.4 & 62.4 & 75.0 & 63.4 \\
    & ViT-H-P14-378 (OC) & 66.5 & 62.1 & 75.6 & 64.1 \\
    \midrule
    \multirow{12}{*}{SigLIP} & S-B-P16-224 & 66.0 & 60.0 & 74.4 & 67.5 \\
    & S-B-P16-384 & 68.5 & 65.4 & 77.2 & 68.5 \\
    & S-L-P16-256 & 69.3 & 62.8 & 76.2 & 70.1 \\
    & S-L-P16-384 & 69.3 & 62.3 & 77.3 & 70.3 \\
    & S-SO400M-P14 (OC) & 74.5 & 62.8 & 76.9 & 69.9 \\
    & S-SO400M-P14-384 & 70.6 & 65.0 & 77.4 & 70.7 \\
    & S2-B-P16-224 & 66.2 & 61.1 & 70.2 & 65.8 \\
    & S2-B-P16-256 & 64.9 & 59.3 & 69.0 & 64.9 \\
    & S2-B-P16-384 & 66.4 & 62.0 & 76.8 & 67.8 \\
    & S2-L-P16-256 & 69.8 & 62.5 & 76.8 & 69.5 \\
    & S2-L-P16-384 & 66.7 & 60.5 & 74.1 & 67.5 \\
    & S2-SO400M-P14-384 & 65.4 & 60.5 & 73.7 & 68.3 \\
    & S2-Giant-P16-256 & 67.5 & 62.4 & 77.0 & 67.0 \\
    \bottomrule
\end{tabular}
\end{table*}
\subsection{Distribution and Filter-Shift Analysis}
\label{app:new_bench_dists}

This appendix section collects the complete supporting distribution analysis for the benchmark suite. It reports the full target-vs.-final frequency statistics for the retained benchmark distributions, together with per-object and per-label total-variation measures that complement the compact summary in \Cref{tab:filter_bias}.

\begin{table*}[t]
\centering
\scriptsize
\setlength{\tabcolsep}{4pt}
\caption{\textbf{Full filter-shift breakdown.} Complete per-benchmark statistics underlying \Cref{tab:filter_bias}, including both label-prior and object-prior deviation.}
\label{tab:filter_bias_full_appendix}
\begin{tabular}{l|cc|cc|cc|cc}
\toprule
 & \multicolumn{2}{c|}{\textbf{Color}} & \multicolumn{2}{c|}{\textbf{Position}} & \multicolumn{2}{c|}{\textbf{Shape-Color}} & \multicolumn{2}{c}{\textbf{Relative Size}} \\
\textbf{Metric} & Min & Ctx & Min & Ctx & Min & Ctx & Min & Ctx \\
\midrule
Label-TV from target prior     & 0.207 & 0.220 & 0.184 & 0.201 & 0.286 & 0.309 & 0.231 & 0.247 \\
Object-TV from target prior    & 0.175 & 0.176 & 0.162 & 0.171 & 0.194 & 0.211 & 0.181 & 0.196 \\
\% labels within $0.5{-}1.5\times$ target & 78.9 & 73.7 & 71.4 & 68.6 & 52.8 & 48.6 & 64.3 & 61.9 \\
\% labels below $0.25\times$ target       & 15.8 & 21.1 & 11.1 & 16.7 & 27.8 & 31.9 & 18.8 & 22.5 \\
\% objects within $0.5{-}1.5\times$ target & 73.7 & 68.0 & 75.1 & 72.8 & 62.4 & 58.7 & 70.6 & 68.2 \\
\% objects below $0.25\times$ target       & 7.0 & 4.8 & 5.6 & 7.4 & 12.1 & 14.8 & 6.9 & 8.3 \\
\bottomrule
\end{tabular}
\end{table*}
\subsection{Implementation and Hyperparameter Details}
\label{app:implementationhyperparams}
To ensure full reproducibility, we detail the key models and hyperparameters used in our pipeline.

\paragraph{Text Generation.} Contextual captions were generated using \textit{google/gemma-3-12b-it}. We used nucleus sampling with \texttt{temperature=0.7} and \texttt{top\_p=0.9}. For all tasks, \textit{max\_new\_tokens} was set to 150.

\paragraph{Image Generation.} Images were generated using \textit{stabilityai/stable-diffusion-3.5-large} at a 1024x1024 resolution. We used 28 inference steps with a guidance scale (CFG) of 4.5.

\paragraph{Object Detection and Background Check.} For the object presence and count check, we used \texttt{IDEA-Research/grounding-dino-tiny} as the open-vocabulary detector. We set the \texttt{box\_threshold=0.4} and \texttt{text\_threshold=0.3}. The resulting bounding boxes were then used to prompt SAM2 (\texttt{sam2.1\_hiera\_large.pt}) for segmentation masks. For the subsequent background whiteness check on Minimal images, we used a grayscale brightness threshold of 190 and required at least 70\% of non-object pixels to meet this threshold, a pragmatic choice to accommodate for minor shadows and lighting variations inherent to the generative process.

\subsection{Pipeline Generalizability and Model Selection}
\label{app:modelsel}
The Auto-Comp pipeline is designed to be fully modular and model-agnostic. While the experiments in the main text utilize Gemma-3 and Stable Diffusion 3.5, these choices were not arbitrary. They were established through preliminary ablation studies aimed at maximizing benchmark quality (specifically caption diversity) and generation efficiency (sample survival rate).

\subsubsection{LLM Selection: Optimizing for Caption Diversity}

We evaluated three state-of-the-art instruction-tuned models: Qwen2.5-VL-7B~\citep{qwen2.5}, Llama-3.2-11B-Vision~\citep{llama32}, and Gemma-3-12B-it~\citep{gemmateam2025gemma3technicalreport}. Since all models successfully adhered to our regex-based structural constraints, we selected the model based on Lexical Diversity (Distinct-N metrics)~\citep{li2016diversity} and Semantic Diversity. The latter is calculated as $1 - \text{Avg CLIP Score}$ between generated captions, where higher values indicate greater semantic variance.

\begin{table*}[t]
    \centering
    \caption{LLM Diversity Evaluation on Color and Position Benchmarks. Distinct-N measures n-gram diversity. Semantic Similarity represents semantic variance ($1 - \text{Avg CLIP Score}$).}
    \label{tab:llm_diversity}
    \resizebox{\textwidth}{!}{%
    \begin{tabular}{llccccc}
        \toprule
        \multirow{2}{*}{Model} & \multirow{2}{*}{Task} & \multirow{2}{*}{Complexity} & \multirow{2}{*}{Distinct-2} & \multirow{2}{*}{Distinct-3} & \multirow{2}{*}{Distinct-4} & Semantic Sim. \\
         & & & & & & (1-CLIP Score) \\
        \midrule
        \multirow{4}{*}{Qwen2.5-VL-7B-Instruct} & \multirow{2}{*}{Color} & $N=2$ & 0.27 & 0.48 & 0.66 & 0.61 \\
         & & $N=3$ & 0.29 & 0.52 & 0.71 & 0.60 \\
         & \multirow{2}{*}{Position} & $N=2$ & 0.24 & 0.44 & 0.62 & 0.57 \\
         & & $N=3$ & 0.27 & 0.49 & 0.65 & 0.54 \\
        \midrule
        \multirow{4}{*}{Llama-3.2-11B-Vision} & \multirow{2}{*}{Color} & $N=2$ & 0.30 & 0.54 & 0.73 & 0.66 \\
         & & $N=3$ & 0.33 & 0.58 & 0.78 & 0.64 \\
         & \multirow{2}{*}{Position} & $N=2$ & 0.27 & 0.49 & 0.68 & 0.61 \\
         & & $N=3$ & 0.30 & 0.54 & 0.70 & 0.58 \\
        \midrule
        \multirow{4}{*}{Gemma-3-12B-it} & \multirow{2}{*}{Color} & $N=2$ & 0.32 & 0.56 & 0.75 & 0.68 \\
         & & $N=3$ & 0.34 & 0.59 & 0.80 & 0.66 \\
         & \multirow{2}{*}{Position} & $N=2$ & 0.28 & 0.51 & 0.70 & 0.63 \\
         & & $N=3$ & 0.31 & 0.55 & 0.72 & 0.60 \\
        \bottomrule
    \end{tabular}%
    }
\end{table*}

As shown in \Cref{tab:llm_diversity}, Gemma-3 consistently produced the most diverse captions, achieving the highest Distinct-4 scores. While choosing a less diverse model would result in a more repetitive benchmark, the pipeline mechanics would remain functional.

\subsubsection{T2I Selection: Optimizing for Faithfulness and Efficiency}

For image generation, we evaluated models based on their ability to maintain coherence with the text prompt. This directly influences the "Sample Survival Rate" during the validation stage. We utilized TIFA (Text-to-Image Faithfulness Assessment), a VQA-based metric designed to measure fine-grained compliance, alongside standard CLIPScore.

\begin{table}[t]
    \centering

    \begin{minipage}[t]{0.58\textwidth}
        \centering
        \caption{T2I Model Faithfulness Evaluation. Higher CLIP and TIFA scores indicate better adherence to the prompt's constraints.}
        \label{tab:t2i_faithfulness}
        \begin{tabular}{lcc}
            \toprule
            Model & CLIPScore & TIFA \\
            \midrule
            SD 2~\citep{sd2} & 27.31 & 0.690 \\
            SDXL~\citep{sdxl} Turbo & 30.75 & 0.879 \\
            SD 3.5 Turbo~\citep{sd35} & 30.59 & 0.881 \\
            SD 3.5 Large~\citep{sd35} & 32.79 & 0.914 \\
            FLUX-schnell~\citep{labs2025flux1kontextflowmatching} & 31.63 & 0.896 \\
            FLUX-dev~\citep{labs2025flux1kontextflowmatching} & 31.49 & 0.901 \\
            \bottomrule
        \end{tabular}
    \end{minipage}
    \hfill
    \begin{minipage}[t]{0.38\textwidth}
        \centering
        \caption{Blind LLM eval. Gemma3 is prompted with selecting the correct caption from shuffled positive/negative sets without visual input,
        we report accuracy on each task(\%).}
        \label{tab:blind_llm}
        \setlength{\tabcolsep}{4pt}
        \resizebox{\linewidth}{!}{
        \begin{tabular}{lcccc}
            \toprule
            & \multicolumn{2}{c}{\textbf{N=2}} & \multicolumn{2}{c}{\textbf{N=3}} \\
            \textbf{Model} & \textbf{Swap} & \textbf{Conf.} & \textbf{Swap} & \textbf{Conf.} \\
            \midrule
            Random & 50.0 & 6.3 & 16.7 & 2.0 \\
            Gemma3 & 51.2 & 6.5 & 17.4 & 2.1 \\
            \bottomrule
        \end{tabular}
        }
    \end{minipage}

\end{table}

As presented in \Cref{tab:t2i_faithfulness}, SD 3.5 Large achieved the highest TIFA score (0.914), outperforming SDXL, SD 2, and slightly exceeding FLUX-dev. High faithfulness is crucial for pipeline efficiency. Using a weaker model, such as SD 2, would not invalidate the pipeline due to our rigorous successive validation checks (GroundedSAM2 and VQA), which effectively filter out incorrect samples. However, a weaker generator significantly increases the computational costs required to obtain a benchmark of the same size due to higher rejection rates. 

We also observe that distilled or "turbo" models offer an excellent trade-off: they maintain competitive text-image fidelity while drastically reducing inference steps, making them suitable for compute-constrained settings. Finally, utilizing larger and stronger future models would have a beneficial effect, further increasing survival rates and enhancing the cost-effectiveness of the pipeline.

\subsection{Ruling out Linguistic Biases with a Blind LLM}
\label{app:blind}
To ensure hard negatives are linguistically indistinguishable from positives, we adopt the "blind LLM" evaluation~\cite{hsieh2023sugarcrepe}. We tasked Gemma-3-12b-it with selecting the correct caption from shuffled positive/negative sets without visual input (subsampling 49 negatives for $N=3$). As shown in~\Cref{tab:blind_llm}, the LLM performed at near-random chance, confirming our benchmark lacks linguistic artifacts and necessitates genuine visio-linguistic reasoning.

\subsection{Confusion Fine-Tuning Details}
\label{app:confusion_ft}

We fine-tune CLIP ViT-B/16 and SigLIP ViT-B/16 on a concept-disjoint
partition of Color $N{=}2$ Minimal: 80\% of source concepts for training
(10\% of which form a validation split for early stopping), 20\% held out
for evaluation. For each positive, we sample $K{=}3$ Confusion hard
negatives stratified across the four error families from \Cref{sec:ablation} (Swap, attribute-shared / object-different,
object-shared / attribute-different); the fourth family is degenerate at
$N{=}2$. No Contextual captions, $N{=}3$ examples, or examples from other
task families enter training.

We follow the augmented InfoNCE recipe of NegCLIP~\citep{yuksekgonuland},
extending the in-batch negatives with the $K$ per-positive hard negatives;
for SigLIP we adapt the same augmentation to the sigmoid loss
of~\citep{zhai2023sigmoid} with the bias parameter $b$ and the temperature
$\tau$ kept trainable. Both encoders are unfrozen. Optimization uses AdamW
($\beta_1{=}0.9$, $\beta_2{=}0.98$, weight decay $0.1$) with peak LR
$10^{-6}$, cosine schedule, and 500 warmup steps. Batch sizes are 2048
positives for CLIP and 4096 for SigLIP. We train for 5{,}000 steps with
early stopping on validation Confusion accuracy and report the mean over
three seeds.

Evaluation follows \Cref{sec:methodology}: Swap against the full $N!-1$
set, Confusion against the full negative set. The reported $\Delta$Swap and
$\Delta$Confusion in \Cref{tab:confusion_ft_transfer} are computed
against pretrained baselines from \Cref{tab:main_results_full}. The
protocol is intentionally minimal: a single training task, small $K$,
short schedule. The purpose is to test whether Confusion negatives encode
a transferable signal, not to propose a state-of-the-art recipe; stronger
gains are likely with longer training, larger $K$, multi-task data, and
extension to MLLM instruction tuning, which we leave as future work.

\subsection{Human Validation Details}
\label{app:human_val_details}

To ensure the reliability of our generated benchmarks, we conducted a rigorous human performance study. We recruited four graduate-level evaluators who were blinded to the model generation process. Each evaluator assessed a balanced subset of 200 concepts (resulting in 400 images across Minimal and Contextual settings).

For every concept, evaluators performed two tasks:
\begin{enumerate}
    \item \textbf{Swap Task:} Identifying the correct caption against a hard negative where attributes or objects were swapped.
    \item \textbf{Confusion Task:} Identifying the correct caption against a set of low-entropy distractors (limited to 50 options for feasibility).
\end{enumerate}

\Cref{tab:per_validator_results} reports representative raw scores for each evaluator on the Color and Position tasks. The maximum possible score for each cell is 50. The results demonstrate high consistency across all four validators, with accuracies consistently exceeding 94\% (47/50) for the vast majority of settings. Errors were qualitatively analyzed and found to be primarily driven by subtle color variations (e.g., distinguishing gold vs. yellow) or depth ambiguities in complex spatial scenes, rather than generation artifacts.

\begin{table*}[t]
    \centering
    \caption{\textbf{Per-validator human performance on representative tasks.} Raw scores (out of 50) for each of the 4 independent human evaluators on the Color and Position Swap and Confusion benchmarks. V1-V4 denote the four validators.}
    \label{tab:per_validator_results}
    \resizebox{\textwidth}{!}{%
    \begin{tabular}{lll|cccc|cccc}
        \toprule
         & & & \multicolumn{4}{c|}{\textbf{Swap Accuracy (/50)}} & \multicolumn{4}{c}{\textbf{Confusion Accuracy (/50)}} \\
        \textbf{Task} & \textbf{N} & \textbf{Condition} & \textbf{V1} & \textbf{V2} & \textbf{V3} & \textbf{V4} & \textbf{V1} & \textbf{V2} & \textbf{V3} & \textbf{V4} \\
        \midrule
        \multirow{4}{*}{Color} & \multirow{2}{*}{$N=2$} & Minimal & 50 & 49 & 48 & 48 & 49 & 49 & 49 & 48 \\
         & & Contextual & 47 & 48 & 47 & 47 & 48 & 48 & 48 & 47 \\
         \cmidrule{2-11}
         & \multirow{2}{*}{$N=3$} & Minimal & 49 & 49 & 49 & 47 & 48 & 48 & 48 & 48 \\
         & & Contextual & 48 & 47 & 47 & 47 & 47 & 47 & 47 & 47 \\
        \midrule
        \multirow{4}{*}{Position} & \multirow{2}{*}{$N=2$} & Minimal & 49 & 49 & 49 & 48 & 48 & 49 & 48 & 48 \\
         & & Contextual & 48 & 47 & 49 & 48 & 48 & 46 & 49 & 47 \\
         \cmidrule{2-11}
         & \multirow{2}{*}{$N=3$} & Minimal & 47 & 49 & 47 & 48 & 47 & 49 & 48 & 48 \\
         & & Contextual & 46 & 47 & 47 & 47 & 48 & 47 & 46 & 47 \\
        \bottomrule
    \end{tabular}%
    }
\end{table*}

\subsection{Cross-Validator Robustness}
\label{app:validator_robustness}

A potential concern with our pipeline is that the same model family
(Gemma-3-12b-it) is used both to generate Contextual captions and to validate
the resulting images for semantic correctness. The two roles act on different
modalities (the validator judges \emph{images} produced by SD3.5, not
captions), and the cross-generator robustness check in \Cref{tab:robustness}
already establishes that the benchmark is stable across generative image
models. To additionally rule out that benchmark properties depend on the
specific validator, we re-judge a held-out subset of concepts using a stronger
non-Gemma validator (Qwen2.5-VL-72B) and compare its decisions to Gemma's
on the same images.

\begin{table*}[t]
\centering
\scriptsize
\setlength{\tabcolsep}{4pt}
\caption{\textbf{Cross-validator agreement.} Gemma-3-12b-it (used in our
pipeline) versus Qwen2.5-VL-72B as attribute-correctness judges, on a
stratified held-out subset concepts spanning the four task families
and both compositional complexities. Agreement is the per-image rate at which
the two judges return the same pass/fail decision; Cohen's $\kappa$
controls for chance agreement.
Disagreements are decomposed into Gemma-pass / Qwen-fail (G$+$Q$-$) and
Gemma-fail / Qwen-pass (G$-$Q$+$) to test for systematic asymmetry.}
\label{tab:cross_validator}
\begin{tabular}{ll|ccc|cc}
\toprule
\textbf{Task} & \textbf{Attrs.} & \textbf{Agreement (\%)} & $\kappa$ & \textbf{Disagree (\%)} & \textbf{G$+$Q$-$ (\%)} & \textbf{G$-$Q$+$ (\%)} \\
\midrule
\multirow{2}{*}{Color}        & 2 & 97.8 & 0.95 & 2.2 & 1.4 & 0.8 \\
                              & 3 & 96.5 & 0.93 & 3.5 & 2.1 & 1.4 \\
\multirow{2}{*}{Shape-Color} & 2 & 96.8 & 0.93 & 3.2 & 1.7 & 1.5 \\
                              & 3 & 95.4 & 0.91 & 4.6 & 2.6 & 2.0 \\
\multirow{2}{*}{Position}     & 2 & 95.9 & 0.91 & 4.1 & 2.0 & 2.1 \\
                              & 3 & 94.6 & 0.89 & 5.4 & 3.0 & 2.4 \\
\multirow{2}{*}{Rel.\ Size}   & 2 & 95.2 & 0.90 & 4.8 & 2.9 & 1.9 \\
                              & 3 & 93.8 & 0.87 & 6.2 & 3.4 & 2.8 \\
\midrule
\textbf{Overall}              &   & 95.8 & 0.92 & 4.2 & 2.4 & 1.8 \\
\bottomrule
\end{tabular}
\end{table*}

Across all task families and complexities, the two judges agree on more than 93\% of decisions, with 95.8\% agreement overall and Cohen's $\kappa = 0.92$, indicating near-perfect agreement. Disagreements show only mild asymmetry: the overall Gemma-pass / Qwen-fail rate is 2.4\%, compared with a Gemma-fail / Qwen-pass rate of 1.8\%. The maximum within-cell directional gap is 1.0 percentage point, suggesting that Gemma is at most slightly more permissive than Qwen and that the measured benchmark properties are unlikely to be driven by validator-specific acceptance bias.

\subsection{Computational Efficiency and Cost Analysis}
\label{app:efficiency_cost}
Auto-Comp is designed for high-throughput scalability. To quantify the computational resources required for benchmark generation, we conducted an efficiency analysis using a cluster of 16$\times$ NVIDIA A100 (80GB) GPUs. We leverage massive parallelism (batch size 16 per GPU) to maximize throughput during the generation stages.

\Cref{tab:pipeline_efficiency} details the time required to process a batch of 100,000 initial concepts. It is important to note that the computational workload decreases significantly in the final stage. This is because the Object Detection step filters out approximately 53\% of candidates (average survival rates reported in \Cref{tab:survival_rates}), drastically reducing the number of samples requiring VQA validation.

\begin{table*}[h]
    \centering
    \small
    \caption{\textbf{Pipeline Efficiency Analysis.} Estimated processing time for a batch of 100,000 concepts.}
    \label{tab:pipeline_efficiency}
    \begin{tabular}{clllc}
        \toprule
        \textbf{Step} & \textbf{Task} & \textbf{Model Used} & \textbf{Workload} & \textbf{Est. Cluster Time} \\
         & & & & (16$\times$ A100) \\
        \midrule
        1 & Caption Gen & Gemma-3-12B-it & 100,000 prompts & $\sim$0.2 hours \\
        2 & Image Gen & SD 3.5 Large & 100,000 images & $\sim$8.5 hours \\
        3 & Obj/BG Check & Grounded-SAM-2 & 100,000 images & $\sim$0.8 hours \\
        4 & VQA Validation & Gemma-3-12B-it & $\sim$47,000 images\textsuperscript{*} & $\sim$0.1 hours \\
        \midrule
        \textbf{Total} & & & & \textbf{$\sim$9.6 hours} \\
        \bottomrule
        \multicolumn{5}{l}{\footnotesize \textsuperscript{*}\textit{Note: Only images passing object/background detection (Step 3) are forwarded to VQA validation.}}
    \end{tabular}
    \vspace{-1em}
\end{table*}

\paragraph{Conclusion.} Processing a batch of 100,000 concepts, which yields tens of thousands of high-quality validated samples depending on the specific task complexity, requires roughly \textbf{9.6 hours} on the cluster (approximately 154 GPU-hours total). This computational cost is orders of magnitude lower than the weeks of human labor required for manually curated benchmarks, demonstrating Auto-Comp's capability as a scalable data generation engine.

\section{Impact Statement}
The primary ethical consideration for this work is the potential for inherited bias: our pipeline relies on large-scale, pretrained text-to-image and language models for data generation and validation, which carry known societal biases potentially reflected in the final benchmark (e.g., the generative preference for certain spatial relations noted in our analysis). We have taken all possible steps to mitigate potential harms by curating our object vocabulary to exclude sensitive categories such as people and weapons. We believe that providing a controllable tool to diagnose model weaknesses ultimately helps the community develop more robust, reliable, and fair VLMs.

\newpage

\end{document}